\documentclass[letterpaper]{article} % DO NOT CHANGE THIS
\ifdefined\anonymousversion
\usepackage[submission]{aaai2027}  % DO NOT CHANGE THIS
\else
\usepackage[preprint]{aaai2027}  % DO NOT CHANGE THIS
\fi
\usepackage[hyphens]{url}  % DO NOT CHANGE THIS
\usepackage{graphicx} % DO NOT CHANGE THIS
\usepackage{natbib}  % DO NOT CHANGE THIS AND DO NOT ADD OPTIONS
\usepackage{caption} % DO NOT CHANGE THIS AND DO NOT ADD OPTIONS
\usepackage{booktabs}
\usepackage{multirow}
\usepackage{amsmath}
\usepackage{amssymb}
\usepackage{algorithm}
\usepackage{algpseudocode}

\newif\ifreviewmarkup
\reviewmarkupfalse % Change to \reviewmarkupfalse for the submission PDF.
\ifreviewmarkup
\usepackage[switch]{lineno}
\fi

\title{BASC: Behavior-Aligned Quantization and Pruning for Low-Bit Spiking Neural Networks}

\author{
    Linliang Chen\textsuperscript{\rm 1,*},
    Yan Zhong\textsuperscript{\rm 2,*}\textsuperscript{\textdagger},
    Xin Liu\textsuperscript{\rm 3},
    Sai Li\textsuperscript{\rm 4}\textsuperscript{\textdaggerdbl},
    Wang Kang\textsuperscript{\rm 4}\textsuperscript{\textdaggerdbl}
}

\affiliations{
    \textsuperscript{\rm 1}Hangzhou International Innovation Institute, Beihang University, Hangzhou 311115, China\\
    \textsuperscript{\rm 2}School of Mathematical Sciences, Peking University, Beijing 100871, China\\
    \textsuperscript{\rm 3}School of Electronic and Information Engineering; \textsuperscript{\rm 4}School of Integrated Circuit Science and Engineering\\
    \textsuperscript{\rm 3,\rm 4}Beihang University, Beijing 100191, China\\
    \textsuperscript{*}Equal contribution. \textsuperscript{\textdagger}Project lead. \textsuperscript{\textdaggerdbl}Corresponding authors.\\
    linliang.chen@buaa.edu.cn, zhongyan@stu.pku.edu.cn, \{saili,wkang\}@buaa.edu.cn
}

\begin{document}

\maketitle
\ifreviewmarkup
\linenumbers
\fi

\begin{abstract}
Spiking Neural Networks (SNNs) encode information through binary spikes and compute in an event-driven manner, offering an energy-efficient paradigm for machine intelligence.
However, high-performance SNNs incur substantial memory and timestep-wise computation costs that hinder deployment on resource-constrained devices. Quantization and pruning provide complementary routes to reducing these costs, yet both make their decisions with local criteria that overlook temporal task feedback in quantization and inter-channel dependencies in pruning. Consequently, optimizing either criterion can still yield suboptimal compression performance.
We refer to this discrepancy as criterion-behavior mismatch and propose Behavior-Aligned SNN Compression (BASC), a unified framework with two lightweight modules. For quantization, the scale is applied to synaptic current at every timestep and therefore shifts spike timing. Temporal-Behavior Scale Correction (TSC) makes the scale learnable under a temporal loss, allowing firing behavior to inform scale optimization. For pruning, channel importance depends on how channels jointly drive the membrane potential across the firing threshold. Boundary-Level Inter-Channel Correction (BIC) uses channelwise importance scores for initial selection and inter-channel information to re-evaluate only channels near the pruning threshold.
Extensive experiments on static and neuromorphic benchmarks show that lower-bit BASC models match or outperform higher-bit baselines and retain this accuracy advantage after structured pruning, while further reducing model storage and synaptic operations.

\end{abstract}

\section{Introduction}

Spiking Neural Networks (SNNs) have emerged as a promising route to energy-efficient machine intelligence, owing to their biological plausibility and spike-driven computation~\cite{maass1997networks,wu2018stbp,pfeiffer2018deep,roy2019towards}.
Unlike Artificial Neural Networks (ANNs), which propagate dense real-valued activations, a spiking neuron accumulates its inputs into a membrane potential over discrete timesteps and emits a binary spike only when a firing threshold is crossed.
The binary nature of spikes replaces the computationally intensive multiply-accumulate (MAC) operations of ANNs with low-cost accumulate (AC) operations, while the event-driven nature of firing further restricts these accumulations to the synapses that actually receive a spike, thereby avoiding unnecessary computation and memory access.
Recent neuromorphic platforms, including Loihi 2~\cite{orchard2021loihi2}, DYNAP-SE2~\cite{richter2024dynapse2}, and SpiNNaker2~\cite{huang2024spinnaker2}, support scalable and energy-efficient SNN inference.

\begin{figure*}[t]
\centering
\includegraphics[width=0.90\textwidth]{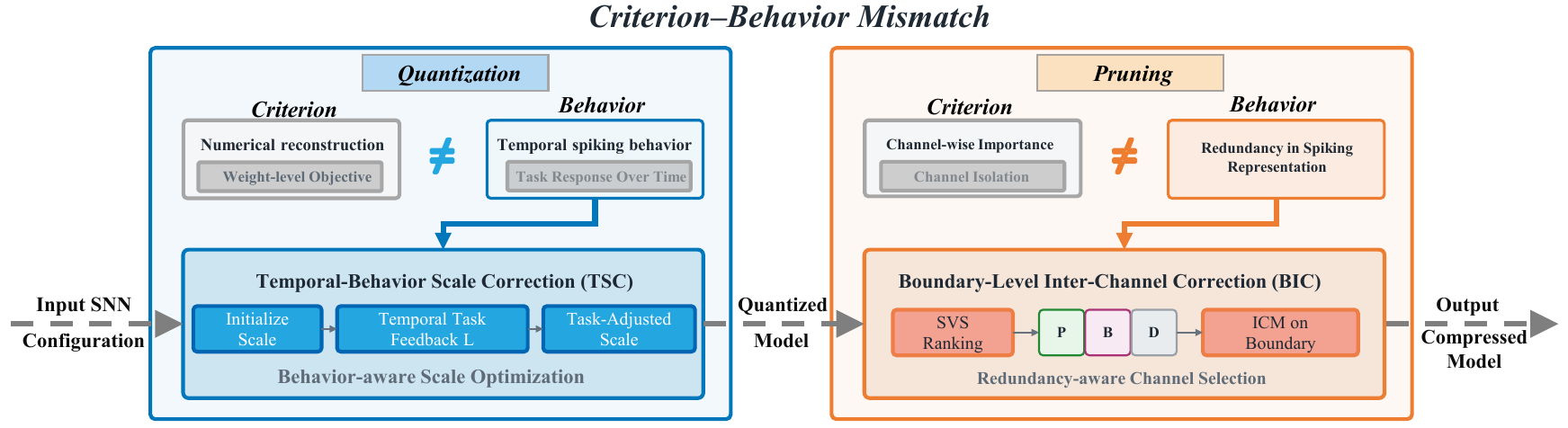}
\caption{Overall pipeline of the proposed joint compression framework.}
\label{fig:overall_pipeline}
\end{figure*}

However, competitive performance on complex tasks often requires scaling up: larger architectures, longer temporal processing, or more elaborate training procedures, all of which increase parameter storage, memory access, and timestep-wise computation\ifdefined\anonymousversion\else~\cite{zhong2026dyn}\fi.
Effective compression is therefore necessary for deployment on resource-constrained devices.
Existing compression methods mainly follow two complementary paths: quantization reduces numerical precision, and pruning removes structural redundancy.

Quantization maps high-precision weights to a small set of discrete values, thereby reducing storage and arithmetic cost~\cite{jacob2018integer,gholami2022survey}.
Structured pruning removes entire channels or filters, allowing the compressed network to maintain a regular shape that is easier to support on hardware.
Although both routes are effective and hardware-friendly, they still face three problems in SNNs.

First, quantization often focuses on local numerical preservation, such as quantization mapping, value range, or reconstruction error.
However, quantized weights change how spike inputs drive neuronal states, further affecting post-quantization spike dynamics.
Second, pruning often selects channels according to the local importance of a single neuron or channel, but the importance of an individual channel cannot fully reflect its redundancy and post-pruning interaction with other channels.
Third, both quantization and pruning rely on local criteria to make compression decisions, while the final performance of an SNN depends on the compressed spatiotemporal dynamics.
Therefore, a locally reasonable criterion does not necessarily lead to optimal compressed SNN performance, and we refer to this discrepancy as \emph{criterion-behavior mismatch}.

To address this mismatch, we propose Behavior-Aligned SNN Compression (BASC), a unified framework for low-bit SNN compression.
Instead of refining the local criteria themselves, BASC gives each of them the information its own scope leaves out: temporal task feedback for quantization, and inter-channel context for pruning.
Figure~\ref{fig:overall_pipeline} illustrates how BASC realizes this alignment for quantization and pruning within a single pipeline.

Specifically, for quantization, we introduce Temporal-Behavior Scale Correction (TSC), which corrects the quantization scale with temporal task feedback.
The quantization scale is no longer fixed only by weight-level numerical criteria, but is optimized jointly with the weights under a loss defined over every timestep, allowing post-quantization spiking behavior to correct quantization decisions.
For pruning, we first use a per-channel importance criterion as an efficient proposal signal, and then use Boundary-Level Inter-Channel Correction (BIC) to revise the resulting ambiguous channel decisions near the pruning threshold, where channels scored in isolation are most likely to hide mutual redundancy.

The contributions of this paper are as follows:
\begin{itemize}
    \item We revisit low-bit SNN compression from the perspective of criterion-behavior mismatch, and propose Behavior-Aligned SNN Compression (BASC), a unified framework designed to make local compression criteria better reflect how compression affects the network.
    \item Under the BASC framework, we introduce TSC for low-bit quantization and BIC for structured pruning, enabling behavior-aware scale optimization and redundancy-aware channel selection.
    \item We conduct systematic experiments on static and neuromorphic benchmark datasets.
    To the best of our knowledge, BASC achieves state-of-the-art accuracy-compression trade-offs under 2/3/4-bit quantization and structured pruning settings, and maintains better performance than existing high-bit baselines at lower bit widths.
\end{itemize}

\section{Related Work}

\subsection{SNN Quantization}
Quantization in SNNs is commonly categorized by how the low-bit model is obtained: conversion from a quantized ANN, or direct low-bit training.
Early conversion-based methods first train a quantized ANN and then map it to a spiking counterpart~\cite{rueckauer2017conversion}, using strategies such as an activation-penalty term~\cite{sorbaro2020optimizing}, binary stochastic activations~\cite{roy2019scaling}, or weight-threshold balancing~\cite{wang2020deep} to limit the loss incurred during conversion, but they typically still suffer from long latency and reduced accuracy~\cite{ding2021optimal,wang2024universal}.
Direct low-bit training instead learns the quantized network end-to-end~\cite{tan2023integerstbp,hu2024bitsnns}.
ALBSNN~\cite{pei2023albsnn} selects layers for binarization using an accuracy-loss estimator based on the discrepancy between binarized and full-precision weights, whereas CBP-QSNN~\cite{yoo2023cbpqsnn} uses a Lagrangian penalty to pull weights toward the nearest quantization level.
To improve low-bit range utilization, Q-SNNs adopt a dual-level quantization scheme for low-bit SNNs~\cite{wei2024qsnn}.
Across these approaches, the quantization decision, whether a per-layer distance, a constraint penalty, or a statistical scale, is fixed by how the weights themselves are distributed rather than by their effect on temporal spike behavior.

\subsection{SNN Pruning}
Pruning in SNNs can be broadly divided into unstructured and structured methods~\cite{vadera2022methods}.
Unstructured pruning removes individual weights and connections for high sparsity, via magnitude- or activity-based criteria~\cite{han2015learning,yin2021energy,shi2024towards} and biologically inspired rewiring~\cite{bellec2017deep,chen2022state}, but yields irregular memory access and therefore depends on dedicated hardware support for acceleration.
Structured pruning removes whole channels or kernels to retain a hardware-friendly shape~\cite{xu2020pruning,hexiao2023structured,li2016pruning}, and prior structured-pruning work differs mainly in its channel-selection criterion: early work scores spatial channel correlations via PCA on membrane potentials~\cite{chowdhury2021spatiotemporal}, while subsequent criteria rank kernels by spike-activity magnitude~\cite{li2024towards}.
These criteria give a stable per-channel ranking, yet the pruning decision is fixed by the score of each channel in isolation rather than by the interaction among the channels retained alongside it.

\subsection{Joint SNN Compression}
Quantization and pruning remove different forms of redundancy, and several studies apply both to compress SNNs further.
\citet{rathi2018stdp} prune insignificant connections under an STDP learning rule with a predefined threshold and then quantize the retained weights.
\citet{deng2023admm} instead cast pruning and quantization as a single constrained optimization problem and solve it with ADMM.
QP-SNN~\cite{wei2025qpsnn} pairs low-bit training with a pruning criterion based on the singular values of spatiotemporal spike activity (SVS), which provides a stable per-channel ranking that is robust to input variation.
These pipelines confirm that quantization and pruning are complementary, but leave their local decision criteria unchanged.

\section{Preliminaries}

\subsection{Spiking Neuron Model}

We use the Leaky Integrate-and-Fire (LIF) neuron~\cite{gerstner2002spiking} and unroll the network over $T$ timesteps with $\mathbf{U}^{l}[0]=\mathbf{0}$.
For layer $l$ at timestep $t$, the pre-reset membrane potential is
\begin{equation}
\tilde{\mathbf{U}}^{l}[t] =
\tau \mathbf{U}^{l}[t-1] + \mathbf{X}^{l}[t],
\quad
\mathbf{X}^{l}[t] = \mathbf{W}^{l}\mathbf{S}^{l-1}[t],
\end{equation}
where $\tau$ is the leak factor, $\mathbf{W}^{l}$ is the synaptic weight, $\mathbf{S}^{l-1}[t]$ is the presynaptic spike, and $\mathbf{X}^{l}[t]$ is the synaptic current.
A spike is emitted through the Heaviside step function $H(\cdot)$ once the potential crosses the firing threshold $\theta$:
\begin{equation}
\mathbf{S}^{l}[t] = H(\tilde{\mathbf{U}}^{l}[t] - \theta).
\end{equation}
Since $H(\cdot)$ is non-differentiable, we train the network with a triangular surrogate gradient of window width $\gamma$,
\begin{equation}
\frac{\partial \mathbf{S}^{l}[t]}{\partial \tilde{\mathbf{U}}^{l}[t]}
\approx
\frac{1}{\gamma^{2}}\max\!\left(\gamma-\bigl|\tilde{\mathbf{U}}^{l}[t]-\theta\bigr|,\,0\right).
\end{equation}
With hard reset, the potential is masked element-wise by the emitted spikes,
\begin{equation}
\mathbf{U}^{l}[t] =
\tilde{\mathbf{U}}^{l}[t]\odot (1-\mathbf{S}^{l}[t]).
\end{equation}

\subsection{Uniform Weight Quantization}

For a bit width $b$, let $Q_p=2^{\,b-1}-1$ denote the number of positive integer grid levels.
Uniform quantization discretizes $\mathbf{W}^{l}$ on a symmetric grid and de-quantizes it as
\begin{equation}
\hat{\mathbf{W}}^{l} =
\frac{\alpha^{l}}{Q_p}
\operatorname{round}\!\left(
Q_p\cdot\mathrm{clip}\!\left(\frac{\mathbf{W}^{l}}{\alpha^{l}}, -1, 1\right)
\right),
\end{equation}
where $\alpha^{l}$ is a layerwise scaling factor, commonly set from a weight statistic such as $\max|\mathbf{W}^{l}|$ and therefore fixed without observing the post-quantization behavior.
The non-differentiable rounding is handled by the straight-through estimator (STE)~\cite{bengio2013estimating}.

\subsection{Structured Channel Pruning}

For a convolutional layer with weight tensor
$\mathbf{W}^{l}\in\mathbb{R}^{C_{out}^{l}\times C_{in}^{l}\times k\times k}$,
where $C_{out}^{l}$ and $C_{in}^{l}$ are the output and input channel counts and $k$ is the kernel size, structured channel pruning removes output channels according to an importance score $g_f$ for each channel $f$, with the layer index omitted for brevity.
Given a pruning ratio $r_l$, the kept channel index set is
\begin{equation}
\mathcal{I}^{l}_{keep}
=
\mathrm{TopK}\left(\{g_f\}_{f=1}^{C_{out}^{l}},
\max\!\left(1,\left\lfloor (1-r_l) C_{out}^{l}\right\rfloor\right) \right),
\end{equation}
where $\mathrm{TopK}(\mathcal{S},K)$ returns the indices of the $K$ largest elements in $\mathcal{S}$, and the $\max(1,\cdot)$ guard prevents a layer from being emptied at high pruning ratios.
Each score $g_f$ is computed from channel $f$ alone, so channels are ranked in isolation, without accounting for interactions among the channels retained together.

\section{Method}

We now instantiate the two corrections that address criterion-behavior mismatch in the compression stages shown in Figure~\ref{fig:overall_pipeline}.
During quantization-aware training, TSC uses temporal task feedback to optimize the layerwise scales.
Afterward, BIC keeps the channelwise proposal and re-evaluates only the decisions near the keep threshold, where isolated scoring is least reliable.

\subsection{Temporal-Behavior Scale Correction}

\begin{figure}[t]
\centering
\includegraphics[width=0.88\columnwidth]{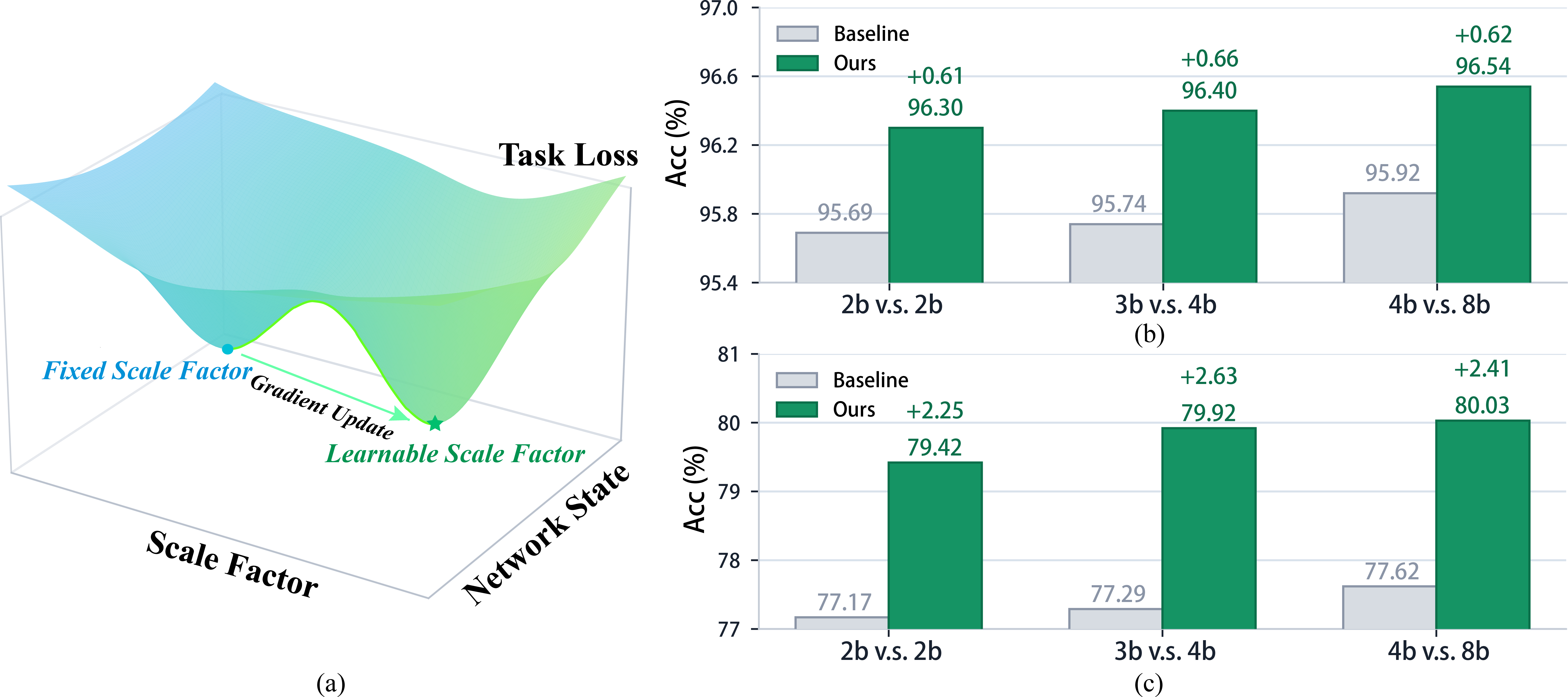}
\caption{Temporal-behavior scale correction. (a) A scale fixed from the weight distribution settles at a suboptimal point of the task loss, whereas a learnable scale is driven toward the task optimum. (b--c) Top-1 accuracy on CIFAR-10 and CIFAR-100 under three bit-width pairings.}
\label{fig:quant_principle}
\end{figure}

\paragraph{Problem analysis.}
Low-bit quantizers commonly set each layerwise scale $\alpha^{l}$ from the weights themselves, either from a statistic of their distribution or by minimizing the quantization error they incur.
Such a criterion is local to the weights, and its optimum does not necessarily coincide with the optimum of the task objective (Figure~\ref{fig:quant_principle}(a)).
This gap is amplified in a spiking network: since $\mathbf{X}^{l}[t]=\hat{\mathbf{W}}^{l}\mathbf{S}^{l-1}[t]$, the scale rescales the synaptic current at every timestep, shifting the membrane-potential trajectory and the firing times it produces.
The quality of a scale is thus determined by the temporal firing pattern it induces, which a weight-level criterion leaves unmeasured.

\paragraph{Learnable temporal scale.}
We therefore make each layerwise scale $\alpha^{l}$ a learnable parameter and optimize it jointly with the weights under the TET temporal loss~\cite{deng2022temporal}. The loss supervises the prediction at each timestep, allowing gradients propagated through the quantized spike sequence to update the scale.
The quantizer is applied to the bounded pre-transform $\mathbf{W}^{l}_{\tanh}=\tanh(\mathbf{W}^{l})$ rather than to $\mathbf{W}^{l}$ directly, which keeps the input to the scale normalization within $(-1,1)$:
\begin{equation}
\hat{\mathbf{W}}^{l}
=\frac{\alpha^{l}}{Q_p}
\operatorname{round}\!\left(
Q_p\cdot\mathrm{clip}\!\left(\frac{\mathbf{W}^{l}_{\tanh}}{\alpha^{l}},-1,1\right)
\right).
\end{equation}
The scale gradient is back-propagated through the quantizer and rescaled by $g=1/\sqrt{N^{l}Q_p}$~\cite{esser2020learned} to balance the update magnitudes of the low-dimensional scale and the high-dimensional weights, where $N^{l}$ is the number of weights in layer $l$.
Our quantizer implementation follows MINT~\cite{yin2024mint}.
The scale is thus corrected by feedback from the post-quantization spiking behavior instead of being fixed by the weight distribution.

\paragraph{Rounding--clipping trade-off.}
To locate the learned scale relative to the error-optimal scale, we freeze the weights and the learned base scales $\alpha^{l}$ after training and sweep a scalar multiplier $m$, evaluating $\alpha^{l}(m)=m\alpha^{l}$ for each candidate without further optimization.
The reconstruction error is minimized at $m_{\mathrm{error}}=0.90$, while validation accuracy peaks at $m_{\mathrm{val}}=1.00$.
Relative to the error-optimal scale, the learned scale lowers the clipping ratio from $2.43\%$ to $0.38\%$ on VGG-16 ($\sim$6.4$\times$) and from $3.62\%$ to $0.46\%$ on ResNet20 ($\sim$7.9$\times$).

\begin{figure}[t]
\centering
\includegraphics[height=0.40\columnwidth,keepaspectratio]{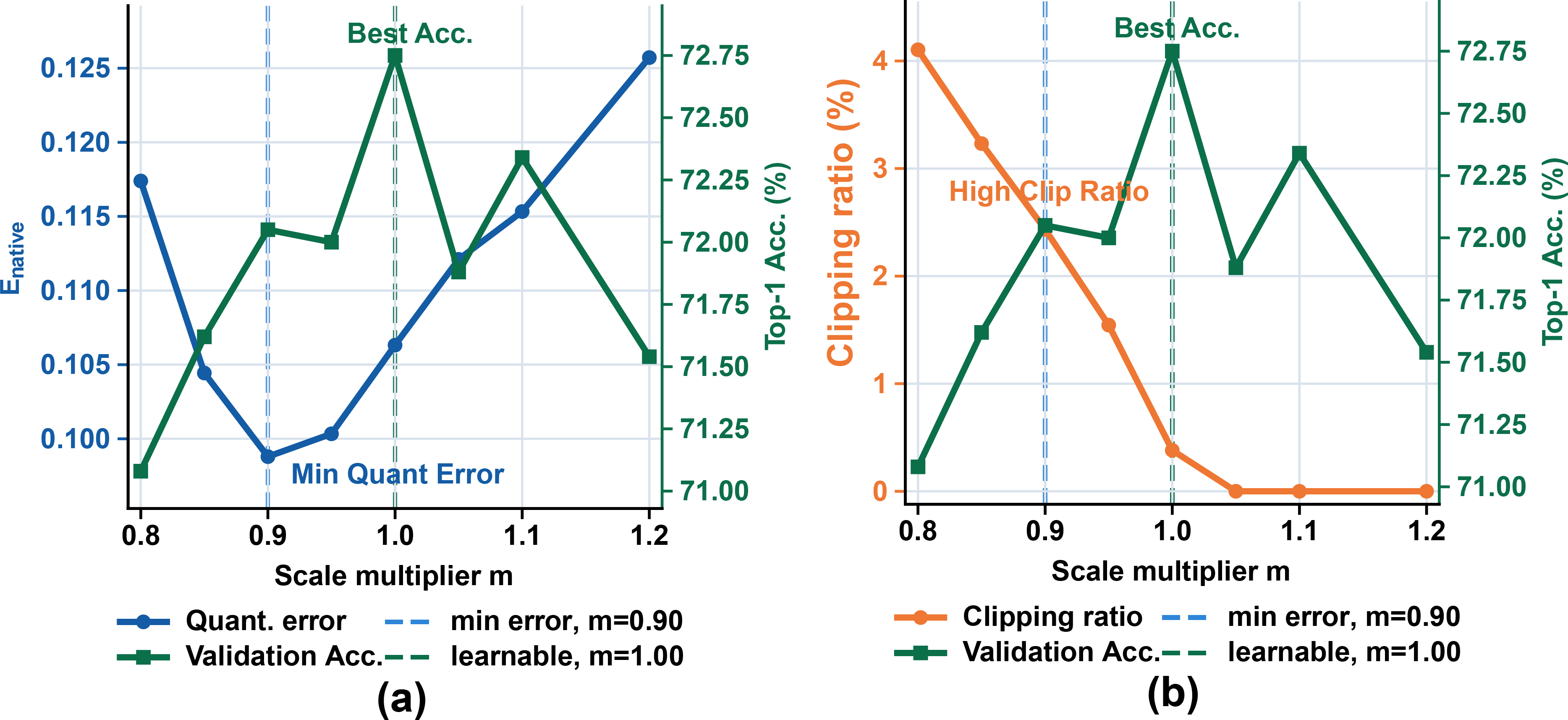}
\caption{Multiplier sweep around the learned scale ($m{=}1.00$).
(a) The reconstruction error is minimized at $m{=}0.90$ while accuracy peaks at $m{=}1.00$: the two optima do not coincide.
(b) The accuracy peak sits at a markedly lower clipping ratio than the error optimum does, showing that the task trades a coarser grid for less clipping.}
\label{fig:quant_sweep}
\end{figure}

To characterize the underlying trade-off, let $v_i=w_{\tanh,i}/\alpha$ and $Q(x)=\mathrm{round}(xQ_p)/Q_p$.
The reconstruction error then decomposes as
\begin{equation}
E(\alpha)^2
=
\sum_{|v_i|\le 1}\alpha^2\bigl(v_i-Q(v_i)\bigr)^2
+
\sum_{|v_i|>1}\bigl(|w_{\tanh,i}|-\alpha\bigr)^2.
\end{equation}
The first term is the rounding error of in-range weights, which grows with $\alpha$ as the grid coarsens; the second is the clipping error of saturated weights, which shrinks with $\alpha$ as fewer weights clip.
The two therefore move in opposite directions, and in our sweep the total error is minimized at a balance point, reached here at $m_{\mathrm{error}}$.
The task-optimal scale $\alpha_{\text{task}}$ is larger than the error-optimal scale, trading a larger rounding error for a smaller clipping error.
Because the clipped weights lie in the tails of the distribution, clipping them removes the largest per-synapse contributions to the membrane drive, and whether such a change flips an output spike depends on the potential jointly produced by all retained channels.
The reduced clipping thus suggests that the tails matter more to firing times than uniform rounding noise does.
Aggregated over all quantized layers, this error is reported as the relative $\ell_2$ norm $E_{\mathrm{native}}=\sqrt{\sum_l E_l(\alpha_l)^2/\sum_l\sum_i(w_i^l)^2}$ in Figure~\ref{fig:quant_sweep}.

The same mismatch appears across additional datasets and three random seeds, and learning the scale instead of fixing it gains $8.25$ points of top-1 accuracy on average across bit widths on CIFAR-100 with VGG-16 (reported in the appendix).
As a result, the corrected scale lets BASC match or exceed the baseline evaluated at the same or a higher bit width (Figure~\ref{fig:quant_principle}(b--c)).

\subsection{Boundary-Level Inter-Channel Correction}

\begin{figure}[t]
\centering
\includegraphics[height=0.40\columnwidth,keepaspectratio]{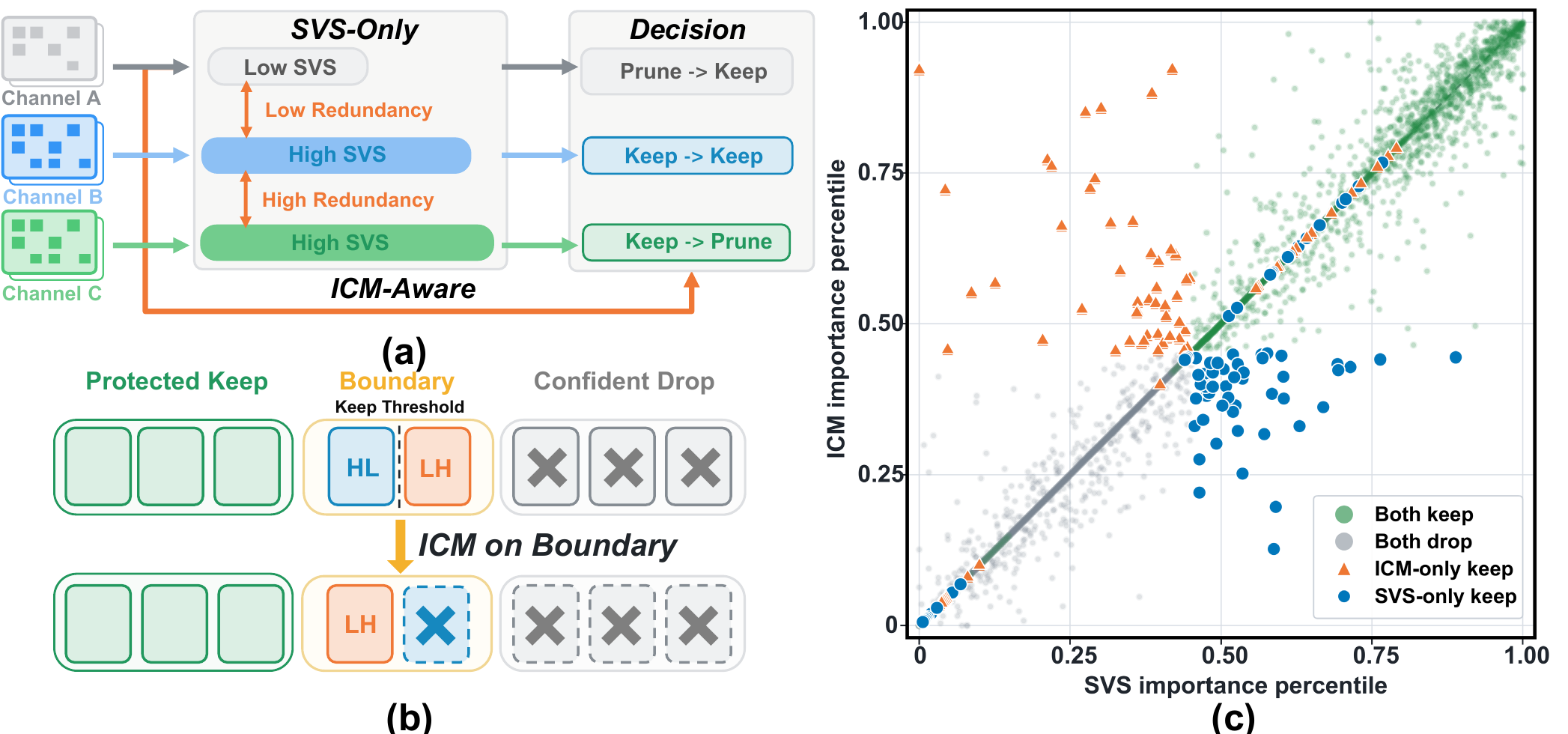}
\caption{Boundary-level inter-channel correction. (a) SVS-only scoring can miss redundancy and yield different keep--prune decisions from ICM-aware evaluation. (b) BIC re-evaluates only boundary channels while preserving confident keep and drop decisions. (c) SVS and ICM agree on most channels, while their disagreements concentrate near the decision boundary.}
\label{fig:prune_boundary}
\end{figure}

\paragraph{Problem analysis.}
Structured pruning criteria commonly score each channel in isolation.
The SVS criterion~\cite{wei2025qpsnn}, for instance, ranks a channel by the effective rank of its own temporally averaged spike map, a score computed from the activity of that channel alone and read off a few calibration batches without labels or back-propagation.
Removing channel $f$, however, shifts the next-layer synaptic current by $\Delta\mathbf{X}^{l+1}[t]=-\hat{\mathbf{W}}^{l+1}_{:,f}\mathbf{S}^{l}_{f}[t]$, which propagates through time as $\Delta\tilde{\mathbf{U}}^{l+1}[t]=\tau\Delta\mathbf{U}^{l+1}[t-1]+\Delta\mathbf{X}^{l+1}[t]$.
Such a perturbation alters the output spikes only where it drives the membrane potential across the firing threshold $\theta$, and that potential is jointly supplied by the channels retained alongside $f$.
The importance of a channel is therefore a property of the retained set rather than of the channel alone: a channel whose isolated score is high may still be redundant with the channels kept beside it (Figure~\ref{fig:prune_boundary}(a)).

\paragraph{Locating the correction.}
Channels ranked far above the pruning threshold are retained and channels ranked far below it are removed almost regardless of the criterion used; the actual decision is determined near the pruning threshold.
Those are also the channels an isolated score separates least well: when two channels score alike on their own, what distinguishes them is how each overlaps with the set retained around it, which a per-channel score does not represent.
We therefore keep the isolated score as a proposal over the full layer and add inter-channel information where the decision is actually made, using the channel-independence score of CHIP~\cite{sui2021chip} as an inter-channel margin (ICM): it measures the nuclear-norm drop of the layer representation after a channel is removed, so a smaller drop marks a channel more redundant with the ones retained around it.

Restricting ICM to the channels near the threshold also keeps it affordable, as each evaluation recomputes the nuclear norm over the full layer activation matrix and is three orders of magnitude slower than scoring the same layers by SVS (Table~\ref{tab:prune_cost}).
The two criteria indeed select the same channels for $95.8\%$ of the decisions and differ almost only near the threshold (Figure~\ref{fig:prune_boundary}(c)), so a full-channel re-evaluation would spend most of its cost where the ranking is not in question; the same pattern holds across bit widths, architectures, and datasets in the appendix.

\paragraph{Protect--boundary split.}
BIC scores every channel by SVS and then re-evaluates with ICM only the channels ranked near the pruning threshold, leaving the rest of the layer to the low-cost SVS score (Figure~\ref{fig:prune_boundary}(b)).
Each channel is first scored by SVS:
\begin{equation}
g_f^{svs}=\mathbb{E}_{n}\big[\operatorname{rank}_{\epsilon}(\bar{\mathbf{A}}_f^{(n)})\big],
\qquad
\bar{\mathbf{A}}_f^{(n)}=\frac{1}{T}\sum_{t=1}^{T}\mathbf{A}_f^{(n)}[t],
\end{equation}
where $\bar{\mathbf{A}}_f^{(n)}$ is the temporally averaged activation map of channel $f$ on input $n$, $\operatorname{rank}_{\epsilon}$ counts singular values above a small threshold $\epsilon$, and $\mathbb{E}_{n}$ averages over inputs.
Let $\pi_l$ denote the permutation that orders channels by descending $g_f^{svs}$. We split this ranking at the keep threshold $\kappa_l$ into three segments, capping the candidate range at $\beta_l=\min(\lceil m\kappa_l\rceil,\,C_{out}^{l})$ so that it never exceeds the width of the layer:
\begin{equation}
\begin{aligned}
\mathcal{P}_l&=\big\{\pi_l(i): i\le\lceil p\kappa_l\rceil\big\},\\
\mathcal{B}_l&=\big\{\pi_l(i): \lceil p\kappa_l\rceil< i\le\beta_l\big\},\\
\mathcal{D}_l&=\big\{\pi_l(i): i>\beta_l\big\},
\end{aligned}
\end{equation}
with $p<1<m$, so that the protect set $\mathcal{P}_l$ is kept and the discarded set $\mathcal{D}_l$ is removed without any inter-channel computation, while the boundary set $\mathcal{B}_l$ brackets the threshold and holds the channels still to be decided.
We set $p=0.95$ and $m=1.25$ and keep both fixed across all experiments.

Stacking the channel activations of input $n$ row by row into $\mathbf{R}^{(n)}\in\mathbb{R}^{C\times(H\times W)}$ and writing $\mathbf{R}^{(n)}_{\ominus f}$ for the same matrix with the row of channel $f$ zeroed, the ICM of a boundary candidate is the drop it causes in the nuclear norm $\lVert\cdot\rVert_{*}$:
\begin{equation}
g_f^{icm}=\mathbb{E}_{n}\big[\lVert\mathbf{R}^{(n)}\rVert_{*}-\lVert\mathbf{R}^{(n)}_{\ominus f}\rVert_{*}\big],
\qquad f\in\mathcal{B}_l.
\end{equation}
Both scores are standardized over the full layer, $z(x_f)=(x_f-\mu)/\sigma$ with $\mu$ and $\sigma$ computed over all $C_{out}^{l}$ channels, and fused as $s_f=z(g_f^{svs})+\lambda\,z(g_f^{icm})$, where $\lambda$ weights the inter-channel term.
The retained set is
\begin{equation}
\begin{gathered}
\mathcal{I}^{l}_{keep}=\mathcal{P}_l\cup\operatorname*{TopK}_{f\in\mathcal{B}_l}\!\big(s_f,\;\kappa_l-|\mathcal{P}_l|\big),\\
\text{s.t.}\quad\big|\mathcal{I}^{l}_{keep}\setminus\mathcal{K}^{svs}_l\big|\le\lfloor\rho\kappa_l\rfloor,
\end{gathered}
\end{equation}
where $\mathcal{K}^{svs}_l$ is the set SVS alone would keep, so the replacement budget $\rho$ caps how many of its choices ICM may displace, and $m>1$ ensures that $\mathcal{B}_l$ holds at least the $\kappa_l-|\mathcal{P}_l|$ channels still to be filled.

\begin{table}[tb]
\centering
\footnotesize
\setlength{\tabcolsep}{3pt}
\begin{tabular}{lccc}
\toprule
Scoring scheme & ICM evals & Rel.\ evals & Wall time (s) \\
\midrule
SVS only & 0 & 0 & 0.88 \\
Full-channel CHIP-CI & 3,712 & 1.00 & 2,995 \\
Boundary-only (ours) & \textbf{572} & \textbf{0.154} & \textbf{480.35} \\
\bottomrule
\end{tabular}
\caption{Scoring cost on CIFAR-10/VGG-16 at 4-bit ($T{=}4$), measured on one RTX 5090 with 6 calibration batches of 256 samples from the same checkpoint. Restricting ICM to the boundary is $6.24\times$ faster than scoring every channel.}
\label{tab:prune_cost}
\end{table}

\section{Experiments}

We first describe the experimental setup, then report quantization and pruning results against a unified baseline, and finally validate each component through an ablation study and an efficiency evaluation.

\subsection{Experimental Setup}

We evaluated BASC on four static image datasets: CIFAR-10, CIFAR-100~\cite{krizhevsky2009learning}, TinyImageNet~\cite{le2015tiny}, and ImageNet-1K~\cite{deng2009imagenet}.
We also evaluated BASC on the neuromorphic dataset DVS-CIFAR10~\cite{li2017cifar10} and used four standard spiking backbones: VGG-16, ResNet20, ResNet18, and VGGSNN.
The number of timesteps was set to $T{=}2$ for ResNet20, $T{=}4$ for VGG-16 and the ImageNet ResNet18, and $T{=}10$ for VGGSNN.
For ImageNet, we also evaluated several backbone variants and downsampling implementations.
The complete comparison is provided in the appendix.

We evaluated 2/3/4-bit quantization.
All models were trained for $300$ epochs using Adam with a learning rate of $0.002$ and a weight decay of $1\times10^{-5}$.
The scale parameters used a separate learning rate of $2.5\times10^{-4}$.
For structured pruning, we combined the SVS criterion with BIC, setting the singular-value threshold to $\epsilon=10^{-6}$ and using a protected-channel ratio $p=0.95$ and a candidate multiplier $m=1.25$.
Both are kept fixed across every dataset, backbone, and bit width, while the fusion weight and replacement budget are $(\lambda,\rho)=(0.05,0.03)$ for ResNet20 and $(0.10,0.05)$ for VGG-16 and VGGSNN; additional recipe experiments are reported in the appendix.
We reuse the per-module channel ratios of QP-SNN without modification, so each compression level fixes the layer widths and the two methods differ only in which channels are kept.
Unless marked otherwise, QP-SNN numbers come from training it under this same configuration, and we refer to it as the baseline throughout.
Additional implementation details are provided in the appendix.

\subsection{Quantization Results}

Table~\ref{tab:quant_main} compares BASC with the QP-SNN baseline in terms of top-1 accuracy and model size.
BASC outperforms the baseline across the evaluated datasets and bit widths.
On CIFAR-100, the 3-bit BASC ResNet20 achieves 79.92\% top-1 accuracy, compared with 77.29\% for the 4-bit baseline, while reducing the model size from 8.77 MB to 6.64 MB.
BASC also improves accuracy across all evaluated settings on CIFAR-10.

The lower-bit BASC models can also match or outperform higher-bit baselines.
On CIFAR-100, the 2-bit BASC ResNet20 achieves 79.42\% accuracy, compared with 77.62\% for the 8-bit baseline, while using a smaller model.
Similar gains are observed on TinyImageNet and DVS-CIFAR10.
On ImageNet, BASC achieves 63.40\% accuracy at 4-bit precision and 60.78\% at 2-bit precision.
These results suggest that optimizing the layerwise scale with feedback from post-quantization spiking behavior provides a better accuracy--bit-width trade-off than fixing the scale from weight statistics.

\begin{table*}[t]
\centering
\footnotesize
\setlength{\tabcolsep}{10pt}
\renewcommand{\arraystretch}{0.85}
\begin{tabular}{@{}llccccc@{}}
\toprule
\textbf{Dataset} & \textbf{Method} & \textbf{Arch.} & \textbf{Bit} & \textbf{T} & \textbf{Acc.\,(\%)} & \textbf{Size (MB)} \\
\midrule
\multirow[c]{5}{*}[-8pt]{CIFAR-10}
 & Deng et al. [TNNLS 2023]\nocite{deng2023admm} & 7Conv2FC & 32/3 & 8 & 90.19/87.59 & 62.16/5.84 \\
\cmidrule(lr){2-7}
 & Wei et al. [ICLR 2025]\nocite{wei2025qpsnn} & \multirow[c]{2}{*}[-2pt]{ResNet20} & 8/4/2 & \multirow[c]{2}{*}[-2pt]{2} & 95.92/95.74/95.69 & 17.10/8.59/4.33 \\
\cmidrule(lr){2-2}\cmidrule(lr){4-4}\cmidrule(lr){6-7}
 & \textbf{BASC (Ours)} & & \textbf{4/3/2} & & \textbf{96.54/96.40/96.30} & \textbf{8.59/6.46/4.33} \\
\cmidrule(lr){2-7}
 & Wei et al. [ICLR 2025]\nocite{wei2025qpsnn} & \multirow[c]{2}{*}[-2pt]{VGG-16} & 8/4/2 & \multirow[c]{2}{*}[-2pt]{4} & 93.33/93.28/93.01 & 14.77/7.42/3.74 \\
\cmidrule(lr){2-2}\cmidrule(lr){4-4}\cmidrule(lr){6-7}
 & \textbf{BASC (Ours)} & & \textbf{4/3/2} & & \textbf{93.76/93.77/93.41} & \textbf{7.42/5.58/3.74} \\
\cmidrule(lr){1-7}
\multirow[c]{5}{*}[-8pt]{CIFAR-100}
 & Deng et al. [TNNLS 2023]\nocite{deng2023admm} & 7Conv2FC & 3/1 & 8 & 57.83/55.95 & 11.75/5.99 \\
\cmidrule(lr){2-7}
 & Wei et al. [ICLR 2025]\nocite{wei2025qpsnn} & \multirow[c]{2}{*}[-2pt]{ResNet20} & 8/4/2 & \multirow[c]{2}{*}[-2pt]{2} & 77.62/77.29/77.17 & 17.29/8.77/4.51 \\
\cmidrule(lr){2-2}\cmidrule(lr){4-4}\cmidrule(lr){6-7}
 & \textbf{BASC (Ours)} & & \textbf{4/3/2} & & \textbf{80.03/79.92/79.42} & \textbf{8.77/6.64/4.51} \\
\cmidrule(lr){2-7}
 & Wei et al. [ICLR 2025]\nocite{wei2025qpsnn} & \multirow[c]{2}{*}[-2pt]{VGG-16} & 8/4/2 & \multirow[c]{2}{*}[-2pt]{4} & 71.12/70.95/70.65 & 14.95/7.60/3.92 \\
\cmidrule(lr){2-2}\cmidrule(lr){4-4}\cmidrule(lr){6-7}
 & \textbf{BASC (Ours)} & & \textbf{4/3/2} & & \textbf{73.28/73.29/72.98} & \textbf{7.60/5.76/3.92} \\
\cmidrule(lr){1-7}
\multirow[c]{2}{*}[-2pt]{TinyImageNet}
 & Wei et al. [ICLR 2025]\nocite{wei2025qpsnn} & \multirow[c]{2}{*}[-2pt]{VGG-16} & 8/4/2 & \multirow[c]{2}{*}[-2pt]{4} & 58.68/58.57/57.84 & 15.16/7.81/4.13 \\
\cmidrule(lr){2-2}\cmidrule(lr){4-4}\cmidrule(lr){6-7}
 & \textbf{BASC (Ours)} & & \textbf{4/3/2} & & \textbf{60.34/60.56/60.15} & \textbf{7.81/5.97/4.13} \\
\cmidrule(lr){1-7}
\multirow[c]{3}{*}[-4pt]{DVS-CIFAR10}
 & Yoo and Jeong [JETCAS 2023]\nocite{yoo2023cbpqsnn} & VGG-16 & 2 & 16 & 74.70 & 3.69 \\
\cmidrule(lr){2-7}
 & Wei et al. [ICLR 2025]\nocite{wei2025qpsnn} & \multirow[c]{2}{*}[-2pt]{VGGSNN} & 8/4/2 & \multirow[c]{2}{*}[-2pt]{10} & 83.20/83.20/82.10 & 9.26/4.66/2.35 \\
\cmidrule(lr){2-2}\cmidrule(lr){4-4}\cmidrule(lr){6-7}
 & \textbf{BASC (Ours)} & & \textbf{4/3/2} & & \textbf{83.80/83.20/84.10} & \textbf{4.66/3.50/2.35} \\
\cmidrule(lr){1-7}
\multirow[c]{3}{*}[-4pt]{ImageNet}
 & Yoo and Jeong [JETCAS 2023]\nocite{yoo2023cbpqsnn} & ResNet18 & 2 & 4 & 54.34 & 3.32 \\
\cmidrule(lr){2-7}
 & Wei et al. [ICLR 2025]\nocite{wei2025qpsnn} & \multirow[c]{2}{*}[-2pt]{ResNet18} & 8/4 & \multirow[c]{2}{*}[-2pt]{4} & 61.36/58.06 & 13.29/7.71 \\
\cmidrule(lr){2-2}\cmidrule(lr){4-4}\cmidrule(lr){6-7}
 & \textbf{BASC (Ours)} & & \textbf{4/2} & & \textbf{63.40/60.78} & \textbf{8.31/5.56} \\
\bottomrule
\end{tabular}
\caption{Quantization-only top-1 accuracy and model size. Baseline rows are trained under our protocol at each bit width; other prior methods are quoted as published. Bold rows denote BASC.}
\label{tab:quant_main}
\end{table*}

\begin{table}[t]
\centering
\scriptsize

\begin{tabular*}{\columnwidth}{@{\extracolsep{\fill}}lccc@{}}
\toprule
\textbf{Configuration} & \textbf{Acc. (\%)} & \textbf{Compared to} & $\boldsymbol{\Delta}$\textbf{\,Acc. (pp)} \\
\midrule
\textbf{\textit{A.}} Fixed scale + SVS & 74.68 & -- & -- \\
\textbf{\textit{B.}} TSC + SVS & 75.94 & \textbf{\textit{A.}} & +1.26 \\
\textbf{\textit{C.}} Fixed scale + BIC & 74.93 & \textbf{\textit{A.}} & +0.25 \\
\textbf{\textit{D.}} TSC + BIC & \textbf{76.04} & \textbf{\textit{A.}} & \textbf{+1.36} \\
\bottomrule
\end{tabular*}

\vspace{1pt}
{\footnotesize (a) Component ablation}
\vspace{5pt}

\setlength{\tabcolsep}{5.75pt}
\begin{tabular*}{\columnwidth}{@{\extracolsep{\fill}}llccc@{}}
\toprule
\textbf{Dataset / Backbone} & \textbf{Method} & \textbf{Acc. (\%)} & \textbf{Energy (mJ)} & $\boldsymbol{\Delta E}$\textbf{ (\%)} \\
\midrule
\multirow{2}{*}{CIFAR-100 / ResNet20} & QP-SNN & 71.56 & 0.2431 & -- \\
 & \textbf{BASC} & \textbf{73.65} & \textbf{0.1960} & \textbf{-19.35} \\
\midrule
\multirow{2}{*}{DVS-CIFAR10 / VGGSNN} & QP-SNN & 78.40 & 0.2725 & -- \\
 & \textbf{BASC} & \textbf{79.70} & \textbf{0.2641} & \textbf{-3.07} \\
\bottomrule
\end{tabular*}

\vspace{1pt}
{\footnotesize (b) Accuracy--energy comparison}

\caption{Component and efficiency results. (a) Ablation of the two corrections on CIFAR-100/ResNet20 under the matched 4-bit, $T{=}2$ setting. (b) Accuracy--energy comparison against the baseline under matched aggressive 4-bit compression.}
\label{tab:components}
\end{table}

\begin{table*}[t]
\centering
\footnotesize
\setlength{\tabcolsep}{10pt}
\renewcommand{\arraystretch}{0.85}
\begin{tabular}{@{}llccccc@{}}
\toprule
\textbf{Dataset} & \textbf{Method} & \textbf{Arch.} & \textbf{Bit} & \textbf{T} & \textbf{Acc.\,(\%)} & \textbf{Size (MB)} \\
\midrule
\multirow[c]{9}{*}[-8pt]{CIFAR-10}
 & Li et al. [ICML 2024]\nocite{li2024towards} & VGG-16 & 32 & 4 & 91.67/90.26 & 17.32/5.68 \\
\cmidrule(lr){2-7}
 & \multirow{2}{*}{Wei et al. [ICLR 2025]}\nocite{wei2025qpsnn} & \multirow[c]{4}{*}[-2pt]{ResNet20} & \multirow{2}{*}{8/4/2} & \multirow[c]{4}{*}[-2pt]{2} & 95.21/95.18/95.07 & 6.27/3.16/1.61 \\
 & & & & & 93.92/93.77/93.53 & 3.92/1.98/1.02 \\
\cmidrule(lr){2-2}\cmidrule(lr){4-4}\cmidrule(lr){6-7}
 & \multirow{2}{*}{\textbf{BASC (Ours)}} & & \multirow{2}{*}{\textbf{4/3/2}} & & \textbf{95.61/95.53/95.40} & \textbf{3.16/2.38/1.61} \\
 & & & & & \textbf{94.82/94.50/94.27} & \textbf{1.98/1.50/1.02} \\
\cmidrule(lr){2-7}
 & \multirow{2}{*}{Wei et al. [ICLR 2025]}\nocite{wei2025qpsnn} & \multirow[c]{4}{*}[-2pt]{VGG-16} & \multirow{2}{*}{8/4/2} & \multirow[c]{4}{*}[-2pt]{4} & 91.61/91.61/91.22 & 4.28/2.16/1.10 \\
 & & & & & 90.69/90.83/89.93 & 1.45/0.74/0.39 \\
\cmidrule(lr){2-2}\cmidrule(lr){4-4}\cmidrule(lr){6-7}
 & \multirow{2}{*}{\textbf{BASC (Ours)}} & & \multirow{2}{*}{\textbf{4/3/2}} & & \textbf{91.94/91.86/91.46} & \textbf{2.16/1.63/1.10} \\
 & & & & & \textbf{91.16/91.00/90.21} & \textbf{0.74/0.57/0.39} \\
\midrule
\multirow[c]{11}{*}[-8pt]{CIFAR-100}
 & Chowdhury et al. [IJCNN 2021]\nocite{chowdhury2021spatiotemporal} & VGG-11 & 32/5 & 50/30 & 67.80/66.20 & 75.90/25.43 \\
 & Shi et al. [ICLR 2024]\nocite{shi2024towards} & ResNet18 & \multirow[c]{2}{*}{32} & \multirow[c]{2}{*}{4} & 72.34/70.45 & 13.18/7.67 \\
 & Li et al. [ICML 2024]\nocite{li2024towards} & VGG-16 & & & 65.53/64.64 & 14.40/9.48 \\
\cmidrule(lr){2-7}
 & \multirow{2}{*}{Wei et al. [ICLR 2025]}\nocite{wei2025qpsnn} & \multirow[c]{4}{*}[-2pt]{ResNet20} & \multirow{2}{*}{8/4/2} & \multirow[c]{4}{*}[-2pt]{2} & 74.54/74.82/74.55 & 6.45/3.35/1.79 \\
 & & & & & 71.21/71.56/71.05 & 4.10/2.17/1.20 \\
\cmidrule(lr){2-2}\cmidrule(lr){4-4}\cmidrule(lr){6-7}
 & \multirow{2}{*}{\textbf{BASC (Ours)}} & & \multirow{2}{*}{\textbf{4/3/2}} & & \textbf{75.92/75.97/75.46} & \textbf{3.35/2.57/1.79} \\
 & & & & & \textbf{73.65/73.29/72.75} & \textbf{2.17/1.69/1.20} \\
\cmidrule(lr){2-7}
 & \multirow{2}{*}{Wei et al. [ICLR 2025]}\nocite{wei2025qpsnn} & \multirow[c]{4}{*}[-2pt]{VGG-16} & \multirow{2}{*}{8/4/2} & \multirow[c]{4}{*}[-2pt]{4} & 64.87/64.57/63.94 & 2.48/1.35/0.79 \\
 & & & & & 63.49/63.57/62.12 & 1.85/1.04/0.63 \\
\cmidrule(lr){2-2}\cmidrule(lr){4-4}\cmidrule(lr){6-7}
 & \multirow{2}{*}{\textbf{BASC (Ours)}} & & \multirow{2}{*}{\textbf{4/3/2}} & & \textbf{66.13/65.90/64.72} & \textbf{1.35/1.07/0.79} \\
 & & & & & \textbf{65.56/65.00/63.54} & \textbf{1.04/0.83/0.63} \\
\midrule
\multirow[c]{6}{*}[-4pt]{TinyImageNet}
 & Kundu et al. [WACV 2021]\nocite{kundu2021spikethrift} & \multirow[c]{2}{*}{VGG-16} & \multirow{2}{*}{32} & 150 & 52.70 & 24.21 \\
 & Li et al. [ICML 2024]\nocite{li2024towards} & & & 4 & 49.36/49.14 & 27.92/19.76 \\
\cmidrule(lr){2-7}
 & \multirow{2}{*}{Wei et al. [ICLR 2025]}\nocite{wei2025qpsnn} & \multirow[c]{4}{*}[-2pt]{VGG-16} & \multirow{2}{*}{8/4/2} & \multirow[c]{4}{*}[-2pt]{4} & 51.46/51.18/51.06 & 4.67/2.55/1.49 \\
 & & & & & 51.02/50.52/50.72 & 3.44/1.94/1.18 \\
\cmidrule(lr){2-2}\cmidrule(lr){4-4}\cmidrule(lr){6-7}
 & \multirow{2}{*}{\textbf{BASC (Ours)}} & & \multirow{2}{*}{\textbf{4/3/2}} & & \textbf{53.38/53.31/52.75} & \textbf{2.55/2.02/1.49} \\
 & & & & & \textbf{52.72/52.77/53.04} & \textbf{1.94/1.56/1.18} \\
\midrule
\multirow[c]{8}{*}[-4pt]{DVS-CIFAR10}
 & Shi et al. [ICLR 2024]\nocite{shi2024towards} & VGGSNN & 32 & 10 & 81.90/78.30 & 14.08/7.24 \\
 & Li et al. [ICML 2024]\nocite{li2024towards} & 5Conv1FC & 32 & 20 & 73.00/71.90 & 3.92/0.32 \\
\cmidrule(lr){2-7}
 & \multirow{3}{*}{Wei et al. [ICLR 2025]}\nocite{wei2025qpsnn} & \multirow[c]{6}{*}[-2pt]{VGGSNN} & \multirow{3}{*}{8/4/2} & \multirow[c]{6}{*}[-2pt]{10} & 79.50/79.40/79.50 & 1.44/0.74/0.39 \\
 & & & & & 78.60/78.40/78.00 & 0.89/0.46/0.25 \\
 & & & & & 74.40/74.00/73.50 & 0.23/0.13/0.08 \\
\cmidrule(lr){2-2}\cmidrule(lr){4-4}\cmidrule(lr){6-7}
 & \multirow{3}{*}{\textbf{BASC (Ours)}} & & \multirow{3}{*}{\textbf{4/3/2}} & & \textbf{80.40/79.90/80.00} & \textbf{0.74/0.56/0.39} \\
 & & & & & \textbf{79.70/79.60/79.40} & \textbf{0.46/0.35/0.25} \\
 & & & & & \textbf{75.40/76.90/74.50} & \textbf{0.13/0.10/0.08} \\
\midrule
\multirow[c]{2}{*}[-2pt]{ImageNet}
 & Wei et al. [ICLR 2025]\nocite{wei2025qpsnn} & \multirow[c]{2}{*}[-2pt]{ResNet18} & 4 & \multirow[c]{2}{*}[-2pt]{4} & 58.06 & 7.71 \\
\cmidrule(lr){2-2}\cmidrule(lr){4-4}\cmidrule(lr){6-7}
 & \textbf{BASC (Ours)} & & \textbf{4} & & \textbf{59.57} & \textbf{6.37} \\
\bottomrule
\end{tabular}
\caption{Top-1 accuracy and model size after quantization and structured pruning. Grouped rows list the evaluated pruning configurations for each backbone; bold rows denote BASC.}
\label{tab:prune_main}
\end{table*}

\subsection{Pruning Results}

Table~\ref{tab:prune_main} reports accuracy and model size after quantization and structured pruning.
At the standard level, BASC outperforms the baseline across all settings.
The largest margin appears on CIFAR-100 with ResNet20, where the 4-bit BASC model achieves 75.92\% accuracy, compared with 74.54\% for the 8-bit baseline, while reducing the model size from 6.45 MB to 3.35 MB.

At the aggressive level, BASC keeps this lead, with the margin remaining stable or widening relative to the standard level.
On CIFAR-100 with VGG-16, the 4-bit margin grows from 1.56 points at the standard level to 1.99 points at the aggressive level, and the aggressive BASC model at 1.04 MB still exceeds the standard baseline at 1.35 MB.
The same protocol carries to ImageNet, where the pruned 4-bit ResNet18 reaches 59.57\% at 6.37 MB.
The ablation study attributes this margin to both components: TSC supplies a stronger low-bit model for the pruning stage to start from, and BIC recovers channel interaction information that the intra-channel SVS proposal alone cannot capture.

\subsection{Ablation Study}

Table~\ref{tab:components}(a) isolates the two corrections on CIFAR-100 with ResNet20, taking fixed-scale quantization with SVS pruning as the reference configuration.
All four models share the same 4-bit, $T{=}2$ setting and fine-tuning recipe, so each row differs only in which correction is applied.
Replacing the fixed scale with TSC improves top-1 accuracy by 1.26 points, and applying both corrections reaches 76.04\%, a gain of 1.36 points over the reference.

Two further checks support the same conclusion beyond this table.
Learning the quantization scale rather than fixing it improves accuracy by 8.25 points on average across bit widths on CIFAR-100 with VGG-16, and the gain from BIC is positive in all four independently trained seeds ($+0.165\pm0.058$ points).
We also checked the one-shot schedule. Two multi-stage variants reaching the same compression in three steps fall below it by 0.70 and 0.87 points, and the larger drop comes from re-scoring the surviving channels at each step.
A score computed on a thinned layer is thus less reliable than one computed on the full layer, which is why BIC proposes over the intact channel set.
Letting pruned channels return during fine-tuning does not surpass one-shot either.
These studies are reported in the appendix.

\subsection{Efficiency Evaluation}

Table~\ref{tab:components}(b) compares BASC with the baseline in estimated compute energy under matched 4-bit compression, following the accounting of \citet{horowitz2014computing}.
BASC is both more accurate and lower in estimated compute energy on a static and a neuromorphic benchmark, cutting it by 19.35\% on CIFAR-100 with ResNet20 and by 3.07\% on DVS-CIFAR10 with VGGSNN.
Most of the remaining settings we evaluated also improve on both axes, while a few trade one against the other; the complete matrix is provided in the appendix.

\section{Conclusion}

In this paper, we revisit low-bit SNN compression from the perspective of criterion-behavior mismatch, where local proxy criteria, such as the numerical fidelity of quantization and the single-channel importance used in pruning, can diverge from the network responses affected by compression.
Guided by this view, we propose Behavior-Aligned SNN Compression (BASC) with two lightweight modules.
For quantization, TSC calibrates the layerwise quantization scale using the temporal task objective.
For pruning, BIC revises ambiguous keep/drop decisions near the pruning threshold using inter-channel information.
Experiments show consistent quantization gains over the baseline and improved structured pruning across all evaluated settings, at both the standard and aggressive compression levels.
Both corrections are confined to where the local criterion carries the least information: TSC adjusts one scalar per layer, and BIC re-evaluates only the channels near the pruning threshold.
The compressed model therefore keeps the same low-bit, structurally pruned form as the baseline.
By using temporal task feedback for quantization and inter-channel redundancy for pruning, BASC offers a unified account of SNN quantization and structured pruning, and future work will pursue broader validation on additional architectures and tasks.

\clearpage
\bibliography{refs}

\begin{thebibliography}{48}
\providecommand{\natexlab}[1]{#1}

\bibitem[{Bellec et~al.(2017)Bellec, Kappel, Maass, and
  Legenstein}]{bellec2017deep}
Bellec, G.; Kappel, D.; Maass, W.; and Legenstein, R. 2017.
\newblock Deep Rewiring: Training Very Sparse Deep Networks.
\newblock \emph{arXiv preprint arXiv:1711.05136}.

\bibitem[{Bengio, L{\'e}onard, and Courville(2013)}]{bengio2013estimating}
Bengio, Y.; L{\'e}onard, N.; and Courville, A. 2013.
\newblock Estimating or Propagating Gradients Through Stochastic Neurons for
  Conditional Computation.
\newblock \emph{arXiv preprint arXiv:1308.3432}.

\bibitem[{Chen et~al.(2022)Chen, Yu, Fang, Ma, Huang, and Tian}]{chen2022state}
Chen, Y.; Yu, Z.; Fang, W.; Ma, Z.; Huang, T.; and Tian, Y. 2022.
\newblock State Transition of Dendritic Spines Improves Learning of Sparse
  Spiking Neural Networks.
\newblock In \emph{International Conference on Machine Learning (ICML)},
  3701--3715. PMLR.

\bibitem[{Chowdhury, Garg, and Roy(2021)}]{chowdhury2021spatiotemporal}
Chowdhury, S.~S.; Garg, I.; and Roy, K. 2021.
\newblock Spatio-Temporal Pruning and Quantization for Low-Latency Spiking
  Neural Networks.
\newblock In \emph{2021 International Joint Conference on Neural Networks},
  1--9.

\bibitem[{Deng et~al.(2009)Deng, Dong, Socher, Li, Li, and
  Fei-Fei}]{deng2009imagenet}
Deng, J.; Dong, W.; Socher, R.; Li, L.-J.; Li, K.; and Fei-Fei, L. 2009.
\newblock ImageNet: A Large-Scale Hierarchical Image Database.
\newblock In \emph{2009 IEEE Conference on Computer Vision and Pattern
  Recognition}, 248--255. IEEE.

\bibitem[{Deng et~al.(2023)Deng, Wu, Hu, Liang, Li, Hu, Ding, Li, and
  Xie}]{deng2023admm}
Deng, L.; Wu, Y.; Hu, Y.; Liang, L.; Li, G.; Hu, X.; Ding, Y.; Li, P.; and Xie,
  Y. 2023.
\newblock Comprehensive SNN Compression Using ADMM Optimization and Activity
  Regularization.
\newblock \emph{IEEE Transactions on Neural Networks and Learning Systems},
  34(6): 2791--2805.

\bibitem[{Deng et~al.(2022)Deng, Li, Zhang, and Gu}]{deng2022temporal}
Deng, S.; Li, Y.; Zhang, S.; and Gu, S. 2022.
\newblock Temporal Efficient Training of Spiking Neural Network via Gradient
  Re-weighting.
\newblock In \emph{International Conference on Learning Representations}.

\bibitem[{Ding et~al.(2021)Ding, Yu, Tian, and Huang}]{ding2021optimal}
Ding, J.; Yu, Z.; Tian, Y.; and Huang, T. 2021.
\newblock Optimal ANN-SNN Conversion for Fast and Accurate Inference in Deep
  Spiking Neural Networks.
\newblock In \emph{Proceedings of the Thirtieth International Joint Conference
  on Artificial Intelligence}, 2328--2336.

\bibitem[{Esser et~al.(2020)Esser, McKinstry, Bablani, Appuswamy, and
  Modha}]{esser2020learned}
Esser, S.~K.; McKinstry, J.~L.; Bablani, D.; Appuswamy, R.; and Modha, D.~S.
  2020.
\newblock Learned Step Size Quantization.
\newblock In \emph{International Conference on Learning Representations}.

\bibitem[{Fang et~al.(2021)Fang, Yu, Chen, Masquelier, Huang, and
  Tian}]{fang2021deep}
Fang, W.; Yu, Z.; Chen, Y.; Masquelier, T.; Huang, T.; and Tian, Y. 2021.
\newblock Deep Residual Learning in Spiking Neural Networks.
\newblock In \emph{Advances in Neural Information Processing Systems},
  volume~34, 21056--21069.

\bibitem[{Gerstner and Kistler(2002)}]{gerstner2002spiking}
Gerstner, W.; and Kistler, W.~M. 2002.
\newblock \emph{Spiking Neuron Models: Single Neurons, Populations,
  Plasticity}.
\newblock Cambridge University Press.

\bibitem[{Gholami et~al.(2022)Gholami, Kim, Dong, Yao, Mahoney, and
  Keutzer}]{gholami2022survey}
Gholami, A.; Kim, S.; Dong, Z.; Yao, Z.; Mahoney, M.~W.; and Keutzer, K. 2022.
\newblock A Survey of Quantization Methods for Efficient Neural Network
  Inference.
\newblock In \emph{Low-Power Computer Vision}, 291--326. Chapman and Hall/CRC.

\bibitem[{Han et~al.(2015)Han, Pool, Tran, and Dally}]{han2015learning}
Han, S.; Pool, J.; Tran, J.; and Dally, W. 2015.
\newblock Learning Both Weights and Connections for Efficient Neural Network.
\newblock \emph{Advances in Neural Information Processing Systems}, 28.

\bibitem[{He and Xiao(2023)}]{hexiao2023structured}
He, Y.; and Xiao, L. 2023.
\newblock Structured Pruning for Deep Convolutional Neural Networks: A Survey.
\newblock \emph{IEEE Transactions on Pattern Analysis and Machine
  Intelligence}.

\bibitem[{Horowitz(2014)}]{horowitz2014computing}
Horowitz, M. 2014.
\newblock Computing's Energy Problem (and What We Can Do about It).
\newblock In \emph{2014 IEEE International Solid-State Circuits Conference
  Digest of Technical Papers (ISSCC)}, 10--14. IEEE.

\bibitem[{Hu, Zheng, and Pan(2024)}]{hu2024bitsnns}
Hu, Y.; Zheng, Q.; and Pan, G. 2024.
\newblock {BitSNNs}: Revisiting Energy-Efficient Spiking Neural Networks.
\newblock \emph{IEEE Transactions on Cognitive and Developmental Systems}.

\bibitem[{Huang et~al.(2024)Huang, Vogginger, Kelber, Gonzalez, Knobloch, and
  Mayr}]{huang2024spinnaker2}
Huang, J.; Vogginger, B.; Kelber, F.; Gonzalez, H.; Knobloch, K.; and Mayr,
  C.~G. 2024.
\newblock Fast Switching Serial and Parallel Paradigms of {SNN} Inference on
  Multi-Core Heterogeneous Neuromorphic Platform {SpiNNaker2}.
\newblock In \emph{2024 International Conference on Neuromorphic Systems
  (ICONS)}, 117--123.

\bibitem[{Jacob et~al.(2018)Jacob, Kligys, Chen, Zhu, Tang, Howard, Adam, and
  Kalenichenko}]{jacob2018integer}
Jacob, B.; Kligys, S.; Chen, B.; Zhu, M.; Tang, M.; Howard, A.; Adam, H.; and
  Kalenichenko, D. 2018.
\newblock Quantization and Training of Neural Networks for Efficient
  Integer-Arithmetic-Only Inference.
\newblock In \emph{Proceedings of the IEEE Conference on Computer Vision and
  Pattern Recognition}, 2704--2713.

\bibitem[{Krizhevsky, Hinton et~al.(2009)}]{krizhevsky2009learning}
Krizhevsky, A.; Hinton, G.; et~al. 2009.
\newblock Learning Multiple Layers of Features from Tiny Images.

\bibitem[{Kundu et~al.(2021)Kundu, Datta, Pedram, and
  Beerel}]{kundu2021spikethrift}
Kundu, S.; Datta, G.; Pedram, M.; and Beerel, P.~A. 2021.
\newblock Spike-Thrift: Towards Energy-Efficient Deep Spiking Neural Networks
  by Limiting Spiking Activity via Attention-Guided Compression.
\newblock In \emph{2021 IEEE Winter Conference on Applications of Computer
  Vision}, 3953--3962.

\bibitem[{Le and Yang(2015)}]{le2015tiny}
Le, Y.; and Yang, X. 2015.
\newblock Tiny ImageNet Visual Recognition Challenge.
\newblock Technical report, Stanford University.

\bibitem[{Li et~al.(2016)Li, Kadav, Durdanovic, Samet, and
  Graf}]{li2016pruning}
Li, H.; Kadav, A.; Durdanovic, I.; Samet, H.; and Graf, H.~P. 2016.
\newblock Pruning Filters for Efficient ConvNets.
\newblock \emph{arXiv preprint arXiv:1608.08710}.

\bibitem[{Li et~al.(2017)Li, Liu, Ji, Li, and Shi}]{li2017cifar10}
Li, H.; Liu, H.; Ji, X.; Li, G.; and Shi, L. 2017.
\newblock {CIFAR10-DVS}: An Event-Stream Dataset for Object Classification.
\newblock \emph{Frontiers in Neuroscience}, 11: 244131.

\bibitem[{Li et~al.(2024)Li, Xu, Shen, Xu, Chen, and Pan}]{li2024towards}
Li, Y.; Xu, Q.; Shen, J.; Xu, H.; Chen, L.; and Pan, G. 2024.
\newblock Towards Efficient Deep Spiking Neural Networks Construction with
  Spiking Activity Based Pruning.
\newblock In \emph{Proceedings of the 41st International Conference on Machine
  Learning}, volume 235 of \emph{Proceedings of Machine Learning Research},
  29063--29073. PMLR.

\bibitem[{Maass(1997)}]{maass1997networks}
Maass, W. 1997.
\newblock Networks of Spiking Neurons: The Third Generation of Neural Network
  Models.
\newblock \emph{Neural Networks}, 10(9): 1659--1671.

\bibitem[{Orchard et~al.(2021)Orchard, Frady, Rubin, Sanborn, Shrestha, Sommer,
  and Davies}]{orchard2021loihi2}
Orchard, G.; Frady, E.~P.; Rubin, D. B.~D.; Sanborn, S.; Shrestha, S.~B.;
  Sommer, F.~T.; and Davies, M. 2021.
\newblock Efficient Neuromorphic Signal Processing with {Loihi 2}.
\newblock In \emph{2021 IEEE Workshop on Signal Processing Systems (SiPS)},
  254--259.

\bibitem[{Pei et~al.(2023)Pei, Xu, Wu, Liu, and Yang}]{pei2023albsnn}
Pei, Y.; Xu, C.; Wu, Z.; Liu, Y.; and Yang, Y. 2023.
\newblock {ALBSNN}: Ultra-Low Latency Adaptive Local Binary Spiking Neural
  Network with Accuracy Loss Estimator.
\newblock \emph{Frontiers in Neuroscience}, 17: 1225871.

\bibitem[{Pfeiffer and Pfeil(2018)}]{pfeiffer2018deep}
Pfeiffer, M.; and Pfeil, T. 2018.
\newblock Deep Learning with Spiking Neurons: Opportunities and Challenges.
\newblock \emph{Frontiers in Neuroscience}, 12: 774.

\bibitem[{Rathi, Panda, and Roy(2018)}]{rathi2018stdp}
Rathi, N.; Panda, P.; and Roy, K. 2018.
\newblock STDP-Based Pruning of Connections and Weight Quantization in Spiking
  Neural Networks for Energy-Efficient Recognition.
\newblock \emph{IEEE Transactions on Computer-Aided Design of Integrated
  Circuits and Systems}, 38(4): 668--677.

\bibitem[{Richter et~al.(2024)Richter, Wu, Whatley, K{\"o}stinger, Nielsen,
  Qiao, and Indiveri}]{richter2024dynapse2}
Richter, O.; Wu, C.; Whatley, A.~M.; K{\"o}stinger, G.; Nielsen, C.; Qiao, N.;
  and Indiveri, G. 2024.
\newblock {DYNAP-SE2}: A Scalable Multi-Core Dynamic Neuromorphic Asynchronous
  Spiking Neural Network Processor.
\newblock \emph{Neuromorphic Computing and Engineering}, 4(1): 014003.

\bibitem[{Roy, Chakraborty, and Roy(2019)}]{roy2019scaling}
Roy, D.; Chakraborty, I.; and Roy, K. 2019.
\newblock Scaling Deep Spiking Neural Networks with Binary Stochastic
  Activations.
\newblock In \emph{2019 IEEE International Conference on Cognitive Computing
  (ICCC)}, 50--58. IEEE.

\bibitem[{Roy, Jaiswal, and Panda(2019)}]{roy2019towards}
Roy, K.; Jaiswal, A.; and Panda, P. 2019.
\newblock Towards Spike-Based Machine Intelligence with Neuromorphic Computing.
\newblock \emph{Nature}, 575(7784): 607--617.

\bibitem[{Rueckauer et~al.(2017)Rueckauer, Lungu, Hu, Pfeiffer, and
  Liu}]{rueckauer2017conversion}
Rueckauer, B.; Lungu, I.-A.; Hu, Y.; Pfeiffer, M.; and Liu, S.-C. 2017.
\newblock Conversion of Continuous-Valued Deep Networks to Efficient
  Event-Driven Networks for Image Classification.
\newblock \emph{Frontiers in Neuroscience}, 11: 682.

\bibitem[{Shi et~al.(2024)Shi, Ding, Hao, and Yu}]{shi2024towards}
Shi, X.; Ding, J.; Hao, Z.; and Yu, Z. 2024.
\newblock Towards Energy Efficient Spiking Neural Networks: An Unstructured
  Pruning Framework.
\newblock In \emph{International Conference on Learning Representations
  (ICLR)}.

\bibitem[{Sorbaro et~al.(2020)Sorbaro, Liu, Bortone, and
  Sheik}]{sorbaro2020optimizing}
Sorbaro, M.; Liu, Q.; Bortone, M.; and Sheik, S. 2020.
\newblock Optimizing the Energy Consumption of Spiking Neural Networks for
  Neuromorphic Applications.
\newblock \emph{Frontiers in Neuroscience}, 14: 516916.

\bibitem[{Sui et~al.(2021)Sui, Yin, Xie, Phan, Zonouz, and Yuan}]{sui2021chip}
Sui, Y.; Yin, M.; Xie, Y.; Phan, H.; Zonouz, S.~A.; and Yuan, B. 2021.
\newblock {CHIP}: {CH}annel Independence-based Pruning for Compact Neural
  Networks.
\newblock In \emph{Advances in Neural Information Processing Systems},
  volume~34, 24604--24616.

\bibitem[{Tan and Wu(2023)}]{tan2023integerstbp}
Tan, P.-Y.; and Wu, C.-W. 2023.
\newblock A Low-Bitwidth Integer-STBP Algorithm for Efficient Training and
  Inference of Spiking Neural Networks.
\newblock In \emph{Proceedings of the 28th Asia and South Pacific Design
  Automation Conference}, 651--656.

\bibitem[{Vadera and Ameen(2022)}]{vadera2022methods}
Vadera, S.; and Ameen, S. 2022.
\newblock Methods for Pruning Deep Neural Networks.
\newblock \emph{IEEE Access}, 10: 63280--63300.

\bibitem[{Wang et~al.(2024)Wang, Liu, Zhang, Luo, and Qu}]{wang2024universal}
Wang, Y.; Liu, H.; Zhang, M.; Luo, X.; and Qu, H. 2024.
\newblock A Universal ANN-to-SNN Framework for Achieving High Accuracy and Low
  Latency Deep Spiking Neural Networks.
\newblock \emph{Neural Networks}, 174: 106244.

\bibitem[{Wang et~al.(2020)Wang, Xu, Yan, and Tang}]{wang2020deep}
Wang, Y.; Xu, Y.; Yan, R.; and Tang, H. 2020.
\newblock Deep Spiking Neural Networks with Binary Weights for Object
  Recognition.
\newblock \emph{IEEE Transactions on Cognitive and Developmental Systems},
  13(3): 514--523.

\bibitem[{Wei et~al.(2024)Wei, Liang, Belatreche, Xiao, Cao, Ren, Wang, Zhang,
  and Yang}]{wei2024qsnn}
Wei, W.; Liang, Y.; Belatreche, A.; Xiao, Y.; Cao, H.; Ren, Z.; Wang, G.;
  Zhang, M.; and Yang, Y. 2024.
\newblock Q-SNNs: Quantized Spiking Neural Networks.
\newblock In \emph{Proceedings of the 32nd ACM International Conference on
  Multimedia}, 8441--8450.

\bibitem[{Wei et~al.(2025)Wei, Zhang, Zhou, Belatreche, Shan, Liang, Cao,
  Zhang, and Yang}]{wei2025qpsnn}
Wei, W.; Zhang, M.; Zhou, Z.; Belatreche, A.; Shan, Y.; Liang, Y.; Cao, H.;
  Zhang, J.; and Yang, Y. 2025.
\newblock QP-SNN: Quantized and Pruned Spiking Neural Networks.
\newblock In \emph{International Conference on Learning Representations}.

\bibitem[{Wu et~al.(2018)Wu, Deng, Li, Zhu, and Shi}]{wu2018stbp}
Wu, Y.; Deng, L.; Li, G.; Zhu, J.; and Shi, L. 2018.
\newblock Spatio-Temporal Backpropagation for Training High-Performance Spiking
  Neural Networks.
\newblock \emph{Frontiers in Neuroscience}, 12: 331.

\bibitem[{Xu et~al.(2020)Xu, Huang, Chen, and Zhang}]{xu2020pruning}
Xu, S.; Huang, A.; Chen, L.; and Zhang, B. 2020.
\newblock Convolutional Neural Network Pruning: A Survey.
\newblock In \emph{2020 39th Chinese Control Conference (CCC)}, 7458--7463.
  IEEE.

\bibitem[{Yin et~al.(2021)Yin, Lee, Kong, Hartvigsen, and Xie}]{yin2021energy}
Yin, H.; Lee, J.~B.; Kong, X.; Hartvigsen, T.; and Xie, S. 2021.
\newblock Energy-Efficient Models for High-Dimensional Spike Train
  Classification Using Sparse Spiking Neural Networks.
\newblock In \emph{Proceedings of the 27th ACM SIGKDD Conference on Knowledge
  Discovery \& Data Mining}, 2017--2025.

\bibitem[{Yin et~al.(2024)Yin, Li, Moitra, and Panda}]{yin2024mint}
Yin, R.; Li, Y.; Moitra, A.; and Panda, P. 2024.
\newblock MINT: Multiplier-less INTeger Quantization for Energy Efficient
  Spiking Neural Networks.
\newblock In \emph{2024 29th Asia and South Pacific Design Automation
  Conference}, 830--835.

\bibitem[{Yoo and Jeong(2023)}]{yoo2023cbpqsnn}
Yoo, D.; and Jeong, D.~S. 2023.
\newblock CBP-QSNN: Spiking Neural Networks Quantized Using Constrained
  Backpropagation.
\newblock \emph{IEEE Journal on Emerging and Selected Topics in Circuits and
  Systems}, 13(4): 1137--1146.

\bibitem[{Zhong et~al.(2026)Zhong, Zhao, Wang, He, Guo, Zhang, Lu, and
  Leng}]{zhong2026dyn}
Zhong, Y.; Zhao, R.; Wang, C.; He, J.; Guo, Q.; Zhang, J.; Lu, Z.; and Leng, L.
  2026.
\newblock Dyn-SSM: Towards the Efficient Long Sequence Learning via
  Bio-interpretable Dynamics in Spiking State Space Models.
\newblock \emph{IEEE Transactions on Cognitive and Developmental Systems}.

\end{thebibliography}

\clearpage
\onecolumn
\appendix
\section*{Appendix}
\def\mainpaper{}
This appendix is organized as a chain of evidence supporting the design choices made in the main paper, rather than as an independent set of ablations.
Sections~\ref{sec:arch}--\ref{sec:impl} fix the experimental setup: which architecture and hyperparameters produce the numbers reported in the main paper. The next two sections defend the paper's two central design decisions in the order they are introduced in the main text: Section~\ref{sec:svsicm} shows that the boundary-only inter-channel correction used by BIC is justified by a global SVS--ICM agreement pattern that holds across architectures and is robust to several nuisance factors, and Section~\ref{sec:scale} first shows that the scale mismatch motivating TSC persists across checkpoints, training recipes, and seeds, then connects the learned correction to threshold-crossing behavior and temporal task gradients under a fixed-weight intervention. Section~\ref{sec:efficiency} then checks whether these two design choices actually translate into the energy reduction claimed in the main paper, reporting the full accuracy--energy picture rather than only the favorable settings. Finally, Section~\ref{sec:pruning-alt} asks whether a different pruning schedule could have done better, and shows that the simpler one-shot design retained in the main paper is not left on the table by an easy alternative.

\section{Architecture Study}
\label{sec:arch}

Table~\ref{tab:imagenet_arch} presents an additional architectural exploration on ImageNet, examining how different backbone variants (SEW ResNet and Spiking ResNet)~\cite{fang2021deep} and downsampling implementations affect accuracy under the same bit width and timestep.
The 4-bit result reported for BASC on ImageNet in the main paper's quantization results table corresponds to the best configuration found here (SEW ResNet with the Conv+BN+SN downsample).
Size is calculated from each instantiated architecture: ReScaWConv weights~\cite{wei2025qpsnn} use 4-bit packing, while ordinary convolution, linear, normalization, bias, and scale parameters remain in FP32.

\begin{table}[htbp]
\centering
\small
\setlength{\tabcolsep}{5pt}
\renewcommand{\arraystretch}{1.05}
\begin{tabular}{@{}lllcccc@{}}
\toprule
\textbf{Dataset} & \textbf{Backbone} & \textbf{Downsample} & \textbf{Bit} & \textbf{T} & \textbf{Acc.\,(\%)} & \textbf{Size (MB)} \\
\midrule
\multirow{6}{*}{ImageNet}
& \multirow{3}{*}{SEW ResNet} & ReScaWConv+BN+SN & 4 & 4 & 62.81 & 7.71 \\
& & Conv+BN+SN & 4 & 4 & 63.40 & 8.31 \\
& & Conv+BN & 4 & 4 & 63.26 & 8.31 \\
\cmidrule(lr){2-7}
& \multirow{3}{*}{Spiking ResNet} & Conv+BN+SN & 4 & 4 & 61.66 & 8.31 \\
& & Conv+BN & 4 & 4 & 62.02 & 8.31 \\
& & Padding & 4 & 4 & 61.89 & 7.61 \\
\bottomrule
\end{tabular}
\caption{Architecture comparison on ImageNet. Size uses decimal MB under architecture-specific mixed-precision storage.}
\label{tab:imagenet_arch}
\end{table}

Conv+BN+SN attains the highest accuracy in this study (63.40\%), which is why it is the configuration used for the ImageNet result in the main paper; ReScaWConv+BN+SN instead gives the smallest model (7.71 MB) at a 0.59-point accuracy cost, and SEW ResNet outperforms Spiking ResNet under every downsampling choice tested here.

\section{Additional Implementation Details}
\label{sec:impl}

Quantized models are trained for 300 epochs with Adam, a base learning rate of $2\times10^{-3}$, and weight decay $10^{-5}$. The quantizer implementation follows MINT~\cite{yin2024mint}. Scale parameters are initialized from the final-weight statistic and optimized by a separate Adam parameter group with learning rate $2.5\times10^{-4}$ and zero weight decay. Their gradients are rescaled by $1/\sqrt{N^lQ_p}$ following learned step-size quantization~\cite{esser2020learned}. SVS and ICM use six calibration mini-batches without labels or back-propagation. Unless otherwise stated, scoring and fine-tuning use a batch size of 256. DVS-CIFAR10/VGGSNN uses a batch size of 64. Pruned models are fine-tuned for 300 epochs with Adam, an initial learning rate of $10^{-3}$, weight decay $10^{-5}$, and cosine annealing. We keep $p=0.95$ and $m=1.25$ fixed in all main experiments. We use $(\lambda,\rho)=(0.05,0.03)$ for ResNet20 and $(0.10,0.05)$ for VGG-16 and VGGSNN. These settings are held fixed throughout the remaining sections unless a table explicitly varies one of them. The analyses in this document are run on independently re-trained or re-instrumented checkpoints of the corresponding configurations, so individual accuracies may differ slightly from the main-paper tables; the main-paper values are the reference numbers for BASC.

\subsection{Pruning Configurations}
\label{sec:prune-config}

To keep the comparison structural rather than merely nominal, BASC reuses the per-module channel pruning ratios published by QP-SNN~\cite{wei2025qpsnn} without modification, so a BASC model and its baseline at the same compression level have identical layer widths and differ only in which channels are kept. Each compression level is therefore identified by the target parameter count of the corresponding QP-SNN configuration, and we refer to these levels by the short codes used in the tables of this document. Table~\ref{tab:keep_plans} lists the codes, and Table~\ref{tab:prune_ratios} gives the underlying per-stage ratios.

\begin{table}[htbp]
\centering
\small
\setlength{\tabcolsep}{5pt}
\begin{tabular}{@{}llllr@{}}
\toprule
\textbf{Code} & \textbf{Dataset} & \textbf{Backbone} & \textbf{Level} & \textbf{Params} \\
\midrule
V425 & CIFAR-10 & VGG-16 & standard & 4.25M \\
V142 & CIFAR-10 & VGG-16 & aggressive & 1.42M \\
V231 & CIFAR-100 & VGG-16 & standard & 2.31M \\
V168 & CIFAR-100 & VGG-16 & aggressive & 1.68M \\
V465 & TinyImageNet & VGG-16 & standard & 4.65M \\
V343 & TinyImageNet & VGG-16 & aggressive & 3.43M \\
T622 & CIFAR-10 & ResNet20 & standard & 6.22M \\
T387 & CIFAR-10 & ResNet20 & aggressive & 3.87M \\
T627 & CIFAR-100 & ResNet20 & standard & 6.27M \\
T392 & CIFAR-100 & ResNet20 & aggressive & 3.92M \\
D146 & DVS-CIFAR10 & VGGSNN & standard & 1.46M \\
D090 & DVS-CIFAR10 & VGGSNN & aggressive & 0.90M \\
D025 & DVS-CIFAR10 & VGGSNN & extreme & 0.25M \\
\bottomrule
\end{tabular}
\caption{Compression-level codes used throughout this document. Each code names the target parameter count of the QP-SNN configuration whose channel pruning ratios BASC reuses.}
\label{tab:keep_plans}
\end{table}

\begin{table}[htbp]
\centering
\small
\setlength{\tabcolsep}{4pt}
\begin{tabular}{@{}llcc@{}}
\toprule
\textbf{Backbone} & \textbf{Stage (channels)} & \textbf{Standard} & \textbf{Aggressive} \\
\midrule
\multirow{2}{*}{VGG-16 (C10)} & layers 1--9 (64--256) & 0.45 & 0.49 \\
 & layers 11--16 (512) & 0.51 & 0.80 \\
\midrule
\multirow{2}{*}{VGG-16 (C100)} & layers 1--9 (64--256) & 0.45 & 0.45 \\
 & layers 11--16 (512) & 0.70 & 0.78 \\
\midrule
\multirow{2}{*}{VGG-16 (Tiny)} & layers 1--9 (64--256) & 0.45 & 0.45 \\
 & layers 11--16 (512) & 0.51 & 0.62 \\
\midrule
\multirow{3}{*}{ResNet20} & conv0 (64) & 0.10 & 0.10 \\
 & layer1 (128) & 0.30/0.60 & 0.35/0.75 \\
 & layer2--3 (256, 512) & 0.60 & 0.75 \\
\midrule
\multirow{2}{*}{VGGSNN} & layers 1--4 (64--256) & 0.50 & 0.50 \\
 & layers 5--10 (256--512) & 0.70 & 0.80 \\
\bottomrule
\end{tabular}
\caption{Per-stage channel pruning ratios, reproduced from QP-SNN~\cite{wei2025qpsnn}. The last convolutional layer of each backbone is left unpruned to preserve the classification head. For VGGSNN, the extreme level (D025) uses 0.82 and 0.93 for the two stages. Within ResNet20's layer1, the two ratios apply to the first and second convolution of each block.}
\label{tab:prune_ratios}
\end{table}

Because the ratios are fixed externally, they are not tuned in this work; the only quantities BASC chooses at the pruning stage are the boundary hyperparameters $p$, $m$, $\lambda$, and $\rho$ given above.

\subsection{Overall Procedure}
\label{sec:algorithm}

Algorithm~\ref{alg:basc} summarizes the full pipeline. The quantization stage uses the TET objective~\cite{deng2022temporal}, the straight-through estimator (STE)~\cite{bengio2013estimating}, and learned step-size gradient rescaling~\cite{esser2020learned}. The pruning stage combines the SVS proposal~\cite{wei2025qpsnn} with the channel-independence score of CHIP~\cite{sui2021chip}. The two corrections act at different stages and are not jointly optimized: TSC operates during quantization-aware training, and BIC operates once on the resulting quantized model, before fine-tuning.

\begin{algorithm}[htbp]
\caption{Behavior-Aligned SNN Compression (BASC)}
\label{alg:basc}
\begin{algorithmic}[1]
\Require Full-precision SNN $\mathcal{M}$; bit width $b$; per-layer ratios $\{r_l\}$; hyperparameters $p,m,\lambda,\rho$; calibration batches $\mathcal{C}$
\Ensure Quantized and pruned model $\mathcal{M}_{q\&p}$
\Statex \textbf{Stage 1: Quantization with TSC}
\State Initialize each scale $\alpha^l$ from the weight statistic of layer $l$
\For{each training epoch}
  \For{each mini-batch}
    \State $\hat{\mathbf{W}}^l \gets$ quantize $\tanh(\mathbf{W}^l)$ with scale $\alpha^l$ \Comment{main-paper Eq.~(7)}
    \State Unroll $T$ timesteps; accumulate the TET loss $\mathcal{L}$
    \State Update $\mathbf{W}^l$ by $\nabla_{\mathbf{W}^l}\mathcal{L}$ through the STE
    \State Update $\alpha^l$ by $g\,\nabla_{\alpha^l}\mathcal{L}$, \; $g=1/\sqrt{N^lQ_p}$
  \EndFor
\EndFor
\Statex \textbf{Stage 2: Pruning with BIC}
\For{each prunable layer $l$}
  \State $\kappa_l \gets \max(1,\lfloor(1-r_l)C_{out}^l\rfloor)$
  \State Score every channel by SVS on $\mathcal{C}$; rank as $\pi_l$
  \State Form $\mathcal{P}_l$ and $\mathcal{B}_l$ from $\pi_l$ using $p$ and $m$ \Comment{main-paper Eq.~(10)}
  \For{$f\in\mathcal{B}_l$} \Comment{intact layer; no re-scoring}
    \State Evaluate $g_f^{icm}$ on $\mathcal{C}$ \Comment{main-paper Eq.~(11)}
  \EndFor
  \State $s_f \gets z(g_f^{svs})+\lambda z(g_f^{icm})$ for $f\in\mathcal{B}_l$ \Comment{$z(\cdot)$ over the full layer}
  \State Select $\kappa_l-|\mathcal{P}_l|$ boundary channels by $s_f$, admitting at
  \Statex \hskip\algorithmicindent most $\lfloor\rho\kappa_l\rfloor$ that SVS would not have kept \Comment{main-paper Eq.~(12)}
  \State $\mathcal{I}^l_{keep} \gets \mathcal{P}_l\cup$ selected channels
\EndFor
\State Remove all channels outside $\{\mathcal{I}^l_{keep}\}$
\State $\mathcal{M}_{q\&p} \gets$ Finetune the pruned model
\end{algorithmic}
\end{algorithm}

\section{Global SVS--ICM Agreement and Robustness}
\label{sec:svsicm}

With the recipe fixed in Section~\ref{sec:impl}, we turn to the first design decision defended in this document: why the pruning stage can rely on the cheap SVS score almost everywhere and only invoke the more expensive inter-channel evaluation near the keep threshold.

Table~\ref{tab:svs_icm_global} aggregates all 12 pruned VGG-16 layers rather than selecting representative layers. Across 2--4 bits, SVS and full-channel ICM retain 94.36--95.85\% of the same channels, while their global rank correlations remain above 0.986. This high global overlap motivates using SVS for the stable core and reserving ICM for a narrow candidate band around the keep threshold. We use $p=0.95$ and $m=1.25$ as fixed values in all experiments.

\begin{table}[htbp]
\centering
\small
\setlength{\tabcolsep}{4.5pt}
\begin{tabular}{@{}ccccccc@{}}
\toprule
\textbf{Bit} & \textbf{Samples} & \textbf{Channels} & \textbf{Agreement (\%)} & \textbf{Retained overlap (\%)} & \textbf{Jaccard (\%)} & \textbf{Spearman $\rho$} \\
\midrule
2 & 1,536 & 3,712 & 95.26 & 95.32 & 91.06 & 0.9864 \\
3 & 1,536 & 3,712 & 94.29 & 94.36 & 89.33 & 0.9902 \\
4 & 1,536 & 3,712 & 95.80 & 95.85 & 92.03 & 0.9964 \\
\bottomrule
\end{tabular}
\caption{Global SVS--ICM agreement on CIFAR-10/VGG-16 (V425), aggregated over all 12 pruned layers. The same six fixed calibration batches are used for both scores.}
\label{tab:svs_icm_global}
\end{table}

Agreement is stable across bit widths, with the retained-set overlap staying within a 94.36--95.85\% band and showing no monotonic trend from 2 to 4 bits. 

Table~\ref{tab:svs_icm_crossarch} extends this analysis to additional architectures and datasets, reporting rank correlation, decision agreement, retained-set overlap, and boundary capture, defined as the fraction of disagreements located within the candidate band around the pruning threshold.

\begin{table}[htbp]
\centering
\small
\begin{tabular}{@{}lcccc@{}}
\toprule
\textbf{Dataset / Backbone} & \textbf{Spearman} & \textbf{Agreement} & \textbf{Retained ov.} & \textbf{Bound.\ capture} \\
\midrule
CIFAR-10 / ResNet20 & 0.9908 & 95.70\% & 96.43\% & 64.10\% \\
CIFAR-100 / ResNet20 & 0.9692 & 94.19\% & 95.18\% & 63.29\% \\
DVS-CIFAR10 / VGGSNN & 0.9614 & 89.64\% & 84.72\% & 49.14\% \\
\bottomrule
\end{tabular}
\caption{Cross-architecture SVS--ICM agreement, 4-bit setting.}
\label{tab:svs_icm_crossarch}
\end{table}

Across these settings, Spearman correlation remains above 0.96, decision agreement ranges from 89.64\% to 95.70\%, and 49.14--64.10\% of the disagreements fall within the boundary band. These results show that the global rankings remain closely related while a substantial fraction of their different keep--prune decisions occurs near the threshold.

We separately varied the replacement budget $\rho$ on the 4-bit setting. Table~\ref{tab:rho_sweep} reports best and final accuracy over $\rho\in[0,0.20]$.

\begin{table}[htbp]
\centering
\small
\begin{tabular}{@{}ccc@{}}
\toprule
$\rho$ & \textbf{Best Acc. (\%)} & \textbf{Final Acc. (\%)} \\
\midrule
0.00 & 91.94 & 91.60 \\
0.01 & 91.85 & 91.40 \\
0.03 & 91.90 & 91.70 \\
0.05 & 91.59 & 91.40 \\
0.10 & 91.90 & 91.54 \\
0.20 & 91.99 & 91.75 \\
\bottomrule
\end{tabular}
\caption{Replacement-budget check on CIFAR-10/VGG-16/B4 (V425).}
\label{tab:rho_sweep}
\end{table}

Across the tested range, best and final accuracy vary by 0.40 and 0.35 points, respectively, indicating limited sensitivity to $\rho$ in this setting.

We next examine whether the scoring signals are stable across calibration batches. Table~\ref{tab:batch_stability} reports pairwise rank correlations and top-$k$ overlaps over six fixed batches.

\begin{table}[htbp]
\centering
\small
\begin{tabular}{@{}lcc@{}}
\toprule
\textbf{Score} & \textbf{Pairwise Spearman} & \textbf{Top-$k$ overlap (\%)} \\
\midrule
SVS & 0.9929 & 97.74 \\
ICM & 0.9932 & 97.68 \\
\bottomrule
\end{tabular}
\caption{Calibration-batch repeatability on CIFAR-10/VGG-16/B4.}
\label{tab:batch_stability}
\end{table}

Both SVS and ICM exceed 0.99 pairwise rank correlation and 97.6\% top-$k$ overlap, showing high repeatability across the sampled calibration batches. We then test whether the BIC gain persists across independently trained seeds.

Table~\ref{tab:paired_seed} reports paired best-accuracy results for four seeds on CIFAR-100/ResNet20/B4/T627. BIC improves over TSC+SVS in all four runs ($+0.165\pm0.058$ points) while replacing only 2--5 channels per seed (keep-plan Jaccard 99.695--99.878\%), showing that the improvement in this setting comes from a small boundary correction rather than a large-scale channel reordering.

\begin{table}[htbp]
\centering
\small
\begin{tabular}{@{}rccc@{}}
\toprule
\textbf{Seed} & \textbf{TSC+SVS} & \textbf{TSC+BIC} & \textbf{Gain} \\
\midrule
0 & 75.94 & 76.04 & +0.10 \\
1 & 75.99 & 76.16 & +0.17 \\
2 & 76.00 & 76.24 & +0.24 \\
3 & 76.34 & 76.49 & +0.15 \\
\midrule
\textbf{Mean $\pm$ std} & \textbf{76.068 $\pm$ 0.184} & \textbf{76.233 $\pm$ 0.190} & \textbf{+0.165 $\pm$ 0.058} \\
\bottomrule
\end{tabular}
\caption{Paired multi-seed best accuracy, TSC+SVS vs.\ TSC+BIC. Sign consistency 4/4 positive.}
\label{tab:paired_seed}
\end{table}

\section{Additional Scale Analysis}
\label{sec:scale}

Section~\ref{sec:svsicm} defended the pruning-side design choice; we now turn to the quantization-side counterpart, TSC, which is motivated by a systematic gap between the reconstruction-optimal and task-optimal quantization scale.

To test whether the learned scale can be explained by weight reconstruction alone, we freeze each trained checkpoint and multiply all learned scales by a global multiplier $m$. Across 19 checkpoints spanning CIFAR-10, CIFAR-100, TinyImageNet, and DVS-CIFAR10 (three backbones, bit-widths 2--4), the multiplier minimizing the aggregate reconstruction error differs from the multiplier maximizing full-set validation accuracy in every case (Table~\ref{tab:scale_sweep}). The learned point $m{=}1$ attains the highest validation accuracy for 15 of the 19 checkpoints, while the remaining four optima lie at adjacent grid points ($m \in \{0.975, 1.025, 1.050\}$). For seven checkpoints, the minimum reconstruction error occurs at the lower boundary of the evaluated grid; we report only the minimum within the evaluated range and do not extrapolate a continuous optimum.

\begin{table}[htbp]
\centering
\small
\begin{tabular}{@{}lccccc@{}}
\toprule
\textbf{Setting} & $m_{\mathrm{error}}$ & \textbf{Acc.\ at } $m_{\mathrm{error}}$ & $m_{\mathrm{val}}$ & \textbf{Best Acc.} & \textbf{Acc.\ at } $m{=}1$ \\
\midrule
CIFAR-10 / ResNet20 / B2 & 0.800 & 90.45 & 1.025 & 96.22 & 96.19 \\
CIFAR-10 / ResNet20 / B3 & 0.900 & 95.90 & 0.975 & 96.25 & 96.12 \\
CIFAR-10 / ResNet20 / B4 & 0.900 & 96.26 & 1.000 & 96.47 & 96.47 \\
CIFAR-10 / VGG-16 / B2 & 0.800 & 58.11 & 1.000 & 93.41 & 93.41 \\
CIFAR-10 / VGG-16 / B3 & 0.800 & 86.76 & 1.000 & 93.79 & 93.79 \\
CIFAR-10 / VGG-16 / B4 & 0.900 & 93.27 & 1.000 & 93.75 & 93.75 \\
CIFAR-100 / ResNet20 / B2 & 0.800 & 58.73 & 1.000 & 79.51 & 79.51 \\
CIFAR-100 / ResNet20 / B3 & 0.900 & 78.98 & 0.975 & 79.64 & 79.53 \\
CIFAR-100 / ResNet20 / B4 & 0.950 & 79.61 & 1.050 & 79.89 & 79.54 \\
CIFAR-100 / VGG-16 / B2 & 0.800 & 61.13 & 1.000 & 72.89 & 72.89 \\
CIFAR-100 / VGG-16 / B3 & 0.875 & 69.20 & 1.000 & 72.87 & 72.87 \\
CIFAR-100 / VGG-16 / B4 (seed A) & 0.900 & 72.05 & 1.000 & 72.75 & 72.75 \\
CIFAR-100 / VGG-16 / B4 (seed B) & 0.900 & 72.09 & 1.000 & 73.17 & 73.17 \\
TinyImageNet / VGG-16 / B2 & 0.800 & 26.00 & 1.000 & 60.38 & 60.38 \\
TinyImageNet / VGG-16 / B3 & 0.925 & 57.54 & 1.000 & 60.44 & 60.44 \\
TinyImageNet / VGG-16 / B4 & 0.975 & 60.05 & 1.000 & 60.12 & 60.12 \\
DVS-CIFAR10 / VGGSNN / B2 & 0.800 & 75.40 & 1.000 & 83.90 & 83.90 \\
DVS-CIFAR10 / VGGSNN / B3 & 0.950 & 82.50 & 1.000 & 83.30 & 83.30 \\
DVS-CIFAR10 / VGGSNN / B4 & 0.975 & 83.20 & 1.000 & 83.80 & 83.80 \\
\bottomrule
\end{tabular}
\caption{Global learned-scale multiplier sweep across 19 checkpoints. Weights and learned scales are frozen; no retraining.}
\label{tab:scale_sweep}
\end{table}

\paragraph{Temporal-behavior mechanism of TSC.}
The global sweeps establish that reconstruction- and task-optimal scales differ, but the TSC claim further requires the learned displacement to be associated with the temporal behavior omitted by reconstruction. We test this link on three seed-0, 2-bit checkpoints using a fixed-weight scale-path intervention. For every quantization scale group, we find the reconstruction-optimal scale $\boldsymbol{\alpha}_{E}$ while keeping the TSC-trained weights fixed, and compare it with the learned scale $\boldsymbol{\alpha}_{\mathrm{TSC}}$. We evaluate five points on the log-scale path
\begin{equation}
\log \boldsymbol{\alpha}(\tau)
=(1-\tau)\log \boldsymbol{\alpha}_{E}
+\tau\log \boldsymbol{\alpha}_{\mathrm{TSC}},
\qquad
\tau\in\{0,0.25,0.50,0.75,1\}.
\label{eq:tsc_scale_path}
\end{equation}
This construction is diagnostic: TSC does not explicitly solve for $\boldsymbol{\alpha}_{E}$ during training. It instead updates the scale with temporal task gradients propagated through the quantized weights, membrane trajectories, and spike generation at every timestep.

Table~\ref{tab:tsc_path_endpoints} reports the two endpoints. Along all three five-point paths, native reconstruction error increases monotonically, while PrefixCE decreases and prefix margin and final accuracy increase monotonically. PrefixCE and prefix margin are computed from predictions accumulated through each timestep, then averaged over time. The endpoint accuracy gains range from 9.00 to 34.45 points; paired 95\% bootstrap confidence intervals exclude zero in every case. BH-corrected McNemar tests reach $p<10^{-300}$ for both CIFAR-100 checkpoints and $p=7.86\times10^{-17}$ for DVS-CIFAR10. The task improvement therefore occurs while the local numerical criterion becomes strictly worse along the evaluated path.

\begin{table}[htbp]
\centering
\footnotesize
\setlength{\tabcolsep}{3.5pt}
\renewcommand{\arraystretch}{1.08}
\begin{tabular}{@{}lccccc@{}}
\toprule
\textbf{Setting} & $E_{\mathrm{native}}$ & \textbf{PrefixCE} & \textbf{Prefix margin} & \textbf{Acc. (\%)} & $\Delta$\textbf{Acc. [95\% CI]} \\
\midrule
C100/VGG-16/T4 & 0.3183 $\rightarrow$ 0.4133 & 2.203 $\rightarrow$ 1.089 & 0.025 $\rightarrow$ 2.373 & 48.78 $\rightarrow$ 72.53 & +23.75 [22.73, 24.76] \\
C100/ResNet20/T2 & 0.3001 $\rightarrow$ 0.3869 & 2.477 $\rightarrow$ 0.916 & $-$0.449 $\rightarrow$ 5.277 & 45.06 $\rightarrow$ 79.51 & +34.45 [33.42, 35.48] \\
DVS-C10/VGGSNN/T10 & 0.3090 $\rightarrow$ 0.3729 & 1.009 $\rightarrow$ 0.678 & 2.518 $\rightarrow$ 3.950 & 74.90 $\rightarrow$ 83.90 & +9.00 [6.90, 11.10] \\
\bottomrule
\end{tabular}
\caption{Fixed-weight scale-path endpoints for three B2 checkpoints. Each arrow runs from the per-group reconstruction optimum $\boldsymbol{\alpha}_{E}$ to the learned TSC scale $\boldsymbol{\alpha}_{\mathrm{TSC}}$. Accuracy intervals use 20,000 paired sample-level bootstrap draws. C100 and DVS-C10 denote CIFAR-100 and DVS-CIFAR10.}
\label{tab:tsc_path_endpoints}
\end{table}

The intermediate activity follows the same direction (Table~\ref{tab:tsc_activity_trace}). At $\boldsymbol{\alpha}_{E}$, total spike activity is lower than at $\boldsymbol{\alpha}_{\mathrm{TSC}}$ at every timestep. A missed crossing denotes an elementwise event where $\boldsymbol{\alpha}_{\mathrm{TSC}}$ produces a spike but $\boldsymbol{\alpha}_{E}$ does not; a spurious crossing denotes the reverse. Missed crossings outnumber spurious crossings in all three models, and 75.0--88.9\% of affected layers show the same under-firing direction. PrefixCE improves at every timestep. These measurements identify under-firing and missed threshold crossings as the observed mechanism in this fixed-weight intervention; they do not imply that a higher global firing rate is universally preferable.

\begin{table}[htbp]
\centering
\footnotesize
\setlength{\tabcolsep}{5pt}
\renewcommand{\arraystretch}{1.08}
\begin{tabular}{@{}lcccc@{}}
\toprule
\textbf{Setting} & $t_{\mathrm{under}}/T$ & $N_{\mathrm{miss}}/N_{\mathrm{spur}}$ \textbf{(M)} & $L_{\mathrm{under}}/L$ & $A_{\mathrm{TET}}$ \\
\midrule
C100/VGG-16/T4 & 4/4 & 446.2 / 382.6 & 9/12 (75.0\%) & 0.8315 \\
C100/ResNet20/T2 & 2/2 & 1748.0 / 1257.9 & 16/18 (88.9\%) & 0.6617 \\
DVS-C10/VGGSNN/T10 & 10/10 & 172.4 / 126.3 & 6/7 (85.7\%) & 0.7843 \\
\bottomrule
\end{tabular}
\caption{Activity and gradient evidence along the fixed-weight correction from $\boldsymbol{\alpha}_{E}$ to $\boldsymbol{\alpha}_{\mathrm{TSC}}$. Here $t_{\mathrm{under}}$ counts under-firing timesteps, $N_{\mathrm{miss}}/N_{\mathrm{spur}}$ gives missed/spurious crossings in millions, and $L_{\mathrm{under}}$ counts affected layers with lower activity at $\boldsymbol{\alpha}_{E}$. Crossing counts compare the same samples and neuron-time elements at the two endpoints. PrefixCE improves and $A_t$ is positive at every timestep for all three settings.}
\label{tab:tsc_activity_trace}
\end{table}

To test whether temporal task feedback can account for the learned displacement, let $\mathbf{d}=\log\boldsymbol{\alpha}_{\mathrm{TSC}}-\log\boldsymbol{\alpha}_{E}$ and measure
\begin{equation}
A_{\mathrm{TET}}
=\cos\!\left(
-\nabla_{\log\boldsymbol{\alpha}}\mathcal{L}_{\mathrm{TET}}(\boldsymbol{\alpha}_{E}),
\mathbf{d}
\right).
\label{eq:tsc_gradient_alignment}
\end{equation}
The aggregate alignment is positive for all three checkpoints ($0.6617$--$0.8315$), as is every per-timestep alignment $A_t$. A small move from $\tau=0$ to $0.05$ reduces TET loss from 2.2171 to 2.1074 on VGG-16, from 2.5368 to 2.3032 on ResNet20, and from 1.1664 to 1.1209 on VGGSNN. For these three B2 checkpoints, the evidence closes the intended TSC chain: a scale selected by the local reconstruction criterion induces distorted threshold-crossing and spike behavior, while the temporal task gradient points toward the learned scale and a task-aligned activity regime even though reconstruction error increases. Because the intervention uses one seed per checkpoint, we limit this mechanism claim to the evaluated models and do not treat it as a cross-seed causal law.

The multiplier sweep above holds the training recipe fixed; Table~\ref{tab:scale_lr} instead checks whether the scale mismatch is an artifact of that fixed recipe. It records an ImageNet optimization-recipe check used during performance tuning. Because the base learning rate and scale learning rate change together, this table should be interpreted as a recipe comparison rather than a single-variable scale-LR sweep.

\begin{table}[htbp]
\centering
\small
\begin{tabular}{@{}cccc@{}}
\toprule
\textbf{Bit} & \textbf{Base LR} & \textbf{Scale LR} & \textbf{Best Acc. (\%)} \\
\midrule
4 & $2\times10^{-3}$ & $2.5\times10^{-4}$ & 61.96 \\
4 & $1\times10^{-3}$ & $5.0\times10^{-4}$ & \textbf{62.39} \\
2 & $2\times10^{-3}$ & $2.5\times10^{-4}$ & 59.88 \\
2 & $1\times10^{-3}$ & $5.0\times10^{-4}$ & \textbf{60.78} \\
\bottomrule
\end{tabular}
\caption{ImageNet/ResNet18 optimization-recipe check. Both learning rates vary together.}
\label{tab:scale_lr}
\end{table}

Best accuracy differs by only 0.43--0.90 points between the two recipes at each bit width, indicating that the recipe choice has a modest but non-negligible effect that is independent of the scale-mismatch pattern documented in Tables~\ref{tab:scale_sweep} and~\ref{tab:scale_seed}.

The scale mismatch illustrated in the main paper's multiplier-sweep figure is consistent across three random seeds and all 12 quantized VGG-16 layers. In each of the 36 layer--seed observations, the reconstruction-error optimum lies below the learned task-aware scale.

\begin{table}[htbp]
\centering
\small
\begin{tabular}{@{}ccccc@{}}
\toprule
\textbf{Seed} & \textbf{Layers} & \textbf{Mean shift} & \textbf{Min--max shift} & $m_{\mathrm{error}}<1$ \\
\midrule
0 & 12 & 12.08\% & 5.00--40.00\% & Yes \\
1 & 12 & 8.75\% & 5.00--15.00\% & Yes \\
2 & 12 & 8.33\% & 5.00--20.00\% & Yes \\
\bottomrule
\end{tabular}
\caption{Layer-wise scale mismatch on CIFAR-100/VGG-16/B4. The shift is measured relative to the learned scale $m{=}1$.}
\label{tab:scale_seed}
\end{table}

Taken together, Tables~\ref{tab:scale_sweep}--\ref{tab:scale_seed} show that the scale mismatch reported in the main paper is not an artifact of a single checkpoint, optimization recipe, or random seed: the reconstruction-optimal scale is consistently below the task-optimal scale across 19 checkpoints, an alternative ImageNet recipe, and three seeds spanning all 12 quantized VGG-16 layers.

The sweeps above perturb a scale that was already learned. Table~\ref{tab:scale_learn} instead compares learning the scale against keeping it fixed throughout training, with the two runs matched within each bit width on CIFAR-100/VGG-16 ($T{=}4$, seed 0, 300 epochs).

\begin{table}[htbp]
\centering
\small
\begin{tabular}{@{}cccc@{}}
\toprule
\textbf{Bit} & \textbf{Learnable scale} & \textbf{Fixed scale} & \textbf{Gain} \\
\midrule
2 & 72.87 & 69.26 & $+3.61$ \\
3 & 72.96 & 65.11 & $+7.85$ \\
4 & 73.23 & 59.93 & $+13.30$ \\
\bottomrule
\end{tabular}
\caption{Learnable versus fixed quantization scale on CIFAR-100/VGG-16. Top-1 accuracy (\%); runs are matched within each bit width.}
\label{tab:scale_learn}
\end{table}

The gain averages $8.25$ points across the three bit widths. We claim this comparison for CIFAR-100/VGG-16 across bit widths only, and do not extend it to other datasets or backbones.

\section{Efficiency Accounting}
\label{sec:efficiency}

Having established that both design choices are individually justified, we now check whether they jointly translate into the energy reduction reported in the main paper. Compute energy is estimated as $E_{\mathrm{compute}} = 0.9\,\mathrm{pJ}\times N_{\mathrm{SOP}} + 4.6\,\mathrm{pJ}\times N_{\mathrm{MAC}}$ for a 45-nm, 0.9-V process~\cite{horowitz2014computing}, a compute-only reference scenario that excludes memory access, batch normalization, membrane updates, and control flow. We report SOPs/sample and estimated compute energy/sample rather than hardware-measured power, since no accelerator-level measurement was collected. Not all settings favor BASC on both axes; we report the full matrix rather than selecting only favorable points.

Table~\ref{tab:dvs_matrix} reports the full same-bit accuracy--energy matrix on DVS-CIFAR10/VGGSNN. Five of nine settings are Pareto improvements; the remainder show explicit trade-offs.

\begin{table}[htbp]
\centering
\small
\setlength{\tabcolsep}{4pt}
\begin{tabular}{@{}rlrrrrrr@{}}
\toprule
\textbf{Bit} & \textbf{Setting} & \textbf{QP Top-1} & \textbf{BASC Top-1} & $\Delta$\textbf{Acc} & \textbf{QP SOP (M)} & \textbf{BASC SOP (M)} & $\Delta$\textbf{Energy} \\
\midrule
2 & D146 & 79.1 & 79.1 & 0.0 & 279.801 & 269.886 & $-$3.43\% \\
2 & D090 & 80.0 & 78.8 & $-$1.2 & 230.228 & 218.547 & $-$4.58\% \\
2 & D025 & 73.5 & 73.6 & +0.1 & 44.678 & 42.814 & $-$2.73\% \\
3 & D146 & 79.0 & 79.7 & +0.7 & 280.579 & 270.407 & $-$3.50\% \\
3 & D090 & 79.5 & 77.0 & $-$2.5 & 229.530 & 222.302 & $-$3.11\% \\
3 & D025 & 73.4 & 76.9 & +3.5 & 44.489 & 46.024 & +2.26\% \\
4 & D146 & 79.1 & 79.2 & +0.1 & 290.625 & 270.756 & $-$5.54\% \\
4 & D090 & 78.4 & 79.7 & +1.3 & 234.636 & 225.335 & $-$3.07\% \\
4 & D025 & 72.9 & 74.6 & +1.7 & 44.042 & 44.626 & +0.86\% \\
\bottomrule
\end{tabular}
\caption{DVS-CIFAR10/VGGSNN same-bit accuracy--energy matrix. Accuracy values are full-test measurements of the checkpoints instrumented for operation counting, and therefore differ from the best-epoch accuracies reported in the main pruning table; this matrix is used only for checkpoint-level accuracy--energy comparison.}
\label{tab:dvs_matrix}
\end{table}

Of the four non-Pareto settings, two trade a large accuracy gain (+3.5 and +1.7 points, at 3- and 4-bit D025) for a small energy increase (+2.26\% and +0.86\%), while the other two trade a small energy reduction for an accuracy decrease; no setting incurs a simultaneous accuracy and energy loss.

Table~\ref{tab:efficiency_extra} reports two additional comparisons: a cross-bit setting where QP-SNN uses 8-bit and BASC uses 4-bit, and settings without a clean energy win, reported for completeness rather than selected for favorable results.

\begin{table}[htbp]
\centering
\small
\begin{minipage}{0.48\textwidth}
\centering
\setlength{\tabcolsep}{3.5pt}
\begin{tabular}{@{}lrrrr@{}}
\toprule
\textbf{Setting} & \textbf{QP B8} & \textbf{BASC B4} & $\Delta$\textbf{Acc} & $\Delta$\textbf{Energy} \\
\midrule
D146 & 79.5 & 79.2 & $-$0.3 & $-$5.01\% \\
D090 & 78.3 & 79.7 & +1.4 & $-$3.66\% \\
D025 & 74.4 & 74.6 & +0.2 & $-$2.85\% \\
\bottomrule
\end{tabular}
\caption*{(a) Cross-bit: QP-SNN B8 vs.\ BASC B4.}
\end{minipage}
\hfill
\begin{minipage}{0.48\textwidth}
\centering
\setlength{\tabcolsep}{4pt}
\begin{tabular}{@{}lrr@{}}
\toprule
\textbf{Setting} & $\Delta$\textbf{Acc} & $\Delta$\textbf{Energy} \\
\midrule
C10/R20/B4/T627 & +0.29 & +0.82\% \\
C10/R20/B4/T392 & +0.18 & +1.65\% \\
Tiny/VGG/B2 (exact) & +1.87 & +7.99\% \\
Tiny/VGG/B2 (cross-budget) & +2.05 & $-$0.47\% \\
\bottomrule
\end{tabular}
\caption*{(b) Settings without a clean energy win, reported for completeness.}
\end{minipage}
\caption{Additional efficiency comparisons.}
\label{tab:efficiency_extra}
\end{table}

Together with the main paper's efficiency comparison table, Tables~\ref{tab:dvs_matrix}--\ref{tab:efficiency_extra} give the complete accuracy--energy picture: BASC is a Pareto improvement in most settings and never loses on both axes simultaneously, but it does not dominate every individual configuration, which we report rather than omit.

\section{Alternative Pruning Schedules}
\label{sec:pruning-alt}

The previous sections defended the choices made in the main paper and confirmed their payoff; this final section checks whether a different pruning schedule would have done better, examining two independent directions: multi-stage schedules and dynamic channel regrowth.

We compared one-shot pruning with two progressive alternatives under the same CIFAR-100/ResNet20/B4/T392 setting. Progressive A applies a fixed final keep plan over three stages, whereas Progressive B re-scores surviving channels at each stage. Table~\ref{tab:alt_schedule} reports the resulting accuracies; for both progressive variants, best accuracy is restricted to the final full-compression stage. The gap between one-shot and the two progressive variants is small but consistent (0.70 and 0.87 points at best accuracy), with one-shot ranking first under both the best- and final-accuracy criteria. Progressive B is the weaker of the two, indicating that scores computed on an already-thinned network are less reliable than scores computed once on the full one, which is also why BIC proposes over the intact layer.

\begin{table}[htbp]
\centering
\small
\setlength{\tabcolsep}{6pt}
\begin{tabular}{@{}lcc@{}}
\toprule
\textbf{Schedule} & \textbf{Best (\%)} & \textbf{Final (\%)} \\
\midrule
One-shot (ours) & \textbf{73.65} & \textbf{73.24} \\
Progressive A: fixed final plan & 72.95 & 72.33 \\
Progressive B: stage-wise re-scoring & 72.78 & 72.06 \\
\bottomrule
\end{tabular}
\caption{Pruning schedules on CIFAR-100/ResNet20 under matched final compression.}
\label{tab:alt_schedule}
\end{table}

We also explored dynamic regrowth signals on CIFAR-10/VGG-16 at the same 29.97\% channel budget, pairing an alternative pruning criterion with two candidate regrowth signals. As an alternative pruning criterion we use spiking-activity-based pruning (SCA)~\cite{li2024towards}, which scores channel $k$ in layer $l$ by $r_k^l=\frac{1}{NT}\sum_{n=1}^N\sum_{t=1}^T\|H_k^l(t)\|$, the average spike count over $N$ calibration samples and $T$ timesteps. A pruned channel is regrown according to the batch-normalization scale $\gamma$ of its layer, tracked as an exponential moving average $\bar{\gamma}_k^l[e]=\beta\bar{\gamma}_k^l[e-1]+(1-\beta)\gamma_k^l[e]$ at epoch $e$; the highest-$\bar{\gamma}_k^l$ pruned channel is periodically reactivated. Replacing the instantaneous BN-$\gamma$ gradient with this EMA signal improves SCA-based regrowth by 1.29 points. Using SVS as the pruning signal with the same EMA regrowth signal gives a similar result. These exploratory variants are reported for completeness; the main method retains the simpler one-shot schedule.

\begin{table}[htbp]
\centering
\small
\begin{tabular}{@{}lcc@{}}
\toprule
\textbf{Prune / regrow} & \textbf{Epochs} & \textbf{Best (\%)} \\
\midrule
SCA / instant BN-$\gamma$ & 233 & 89.40 \\
SCA / EMA BN-$\gamma$ & 233 & \textbf{90.69} \\
SVS / EMA BN-$\gamma$ & 233 & 90.60 \\
\bottomrule
\end{tabular}
\caption{Exploratory dynamic-regrowth signal comparison on CIFAR-10/VGG-16 under a matched channel budget.}
\label{tab:regrow_signal}
\end{table}

SVS/EMA falls only 0.09 points short of SCA/EMA, showing that the EMA regrowth signal helps regardless of which criterion selects pruning candidates.

Across Sections~\ref{sec:svsicm}--\ref{sec:pruning-alt}, the evidence supports the two central design choices in the main paper as broad, robust patterns rather than single-setting artifacts, confirms that the resulting efficiency gains hold on the majority of evaluated configurations, and shows that the simplest exploratory alternatives to the pruning schedule do not outperform the design retained in the main paper.

\end{document}